\documentclass[manuscript,screen,nonacm]{acmart}
\AtBeginDocument{%
  \providecommand\BibTeX{{%
    \normalfont B\kern-0.5em{\scshape i\kern-0.25em b}\kern-0.8em\TeX}}}

\usepackage{multirow}

\begin{document}

% \title{Foundation Multimodal Models: A Review of Architectures, Challenges and Opportunities}
\title{Generalist Multimodal AI: A Review of Architectures, Challenges and Opportunities}

% % \authornote{Both authors contributed equally to this research.}
\author{Sai Munikoti}
\affiliation{%
   \institution{Pacific Northwest National Lab}
   \city{Richland}
   \state{WA}
   \country{USA}}
\email{sai.munikoti@pnnl.gov}

\author{Ian Stewart}
\affiliation{%
   \institution{Pacific Northwest National Lab}
   \city{Richland}
   \state{WA}
   \country{USA}}
\email{ian.stewart@pnnl.gov}

\author{Sameera Horawalavithana}
\affiliation{%
   \institution{Pacific Northwest National Lab}
   \city{Richland}
   \state{WA}
   \country{USA}}
\email{yasanka.horawalavithana@pnnl.gov}

\author{Henry Kvinge}
\affiliation{%
   \institution{Pacific Northwest National Lab}
   \city{Richland}
   \state{WA}
   \country{USA}}
\email{henry.kvinge@pnnl.gov}

\author{Tegan Emerson}
\affiliation{%
   \institution{Pacific Northwest National Lab}
   \city{Richland}
   \state{WA}
   \country{USA}}
\email{tegan.emerson@pnnl.gov}

\author{Sandra E Thompson}
\affiliation{%
   \institution{Pacific Northwest National Lab}
   \city{Richland}
   \state{WA}
   \country{USA}}
\email{sandy.thompson@pnnl.gov}

\author{Karl Pazdernik}
\affiliation{%
   \institution{Pacific Northwest National Lab}
   \city{Richland}
   \state{WA}
   \country{USA}}
\email{karl.pazdernik@pnnl.gov}

% \affiliation{%

% \affiliation{%

\renewcommand{\shortauthors}{Munikoti, et al.}

\begin{abstract} 
% Multimodal models, which learn from two or more modalities, are 
Multimodal models are expected to be a critical component to future advances in artificial intelligence. 
This field is starting to grow rapidly with a surge of new design elements motivated by the success of foundation models in natural language processing (NLP) and vision. 
It is widely hoped that further extending the foundation models to multiple modalities (e.g., text, image, video, sensor, time series, graph, etc.) will ultimately lead to generalist multimodal models, i.e. one model across different data modalities and tasks.  
However, there is little research that systematically analyzes recent multimodal models (particularly the ones that work beyond text and vision) with respect to the underling architecture proposed. 
Therefore, this work provides a fresh perspective on 
% foundation multimodal models (FMMs) 
generalist multimodal models (GMMs)
via a novel architecture and training configuration specific taxonomy. 
This includes factors such as {\em Unifiability}, {\em Modularity}, and {\em Adaptability} that are pertinent and essential to the wide adoption and application of GMMs. 
The review further highlights key challenges and prospects for the field and guide the researchers into the new advancements.
\end{abstract}





\maketitle

\section{Introduction}
Multimodal models are deep learning models that can learn across more than one data modality. It is conjectured that such models may be a necessary step towards artificial general intelligence; therefore, the machine learning community's interest in them is rapidly increasing. The ultimate goal of multimodal learning is to develop a single  model that can perform (or easily adapt to perform) diverse multimodal tasks. A simple example of multimodality is a visual language model that performs both unimodal tasks (e.g., text generation, image classification) and cross-modal tasks (e.g., text-to-image retrieval or image captioning), the latter of which requires both in-context and joint learning across modalities \cite{lu2022unified}.

Multimodal research has been actively pursued throughout much of the history of machine learning \cite{akbari2021vatt,jaegle2021perceiver,jaegle2021perceiverio,huo2021wenlan,singh2022flava,wang2022omnivl,li2022mplug,huang2023language}. However, this research is skewed toward cross-modal learning and a limited range of modalities (text and image). As such, the design elements of model architectures are not sufficient to facilitate a smooth transition to contemporary research of more generalist models. For example, unlike conventional machine learning (ML) models, foundation models are trained to reconstruct massive sets of (generally unlabeled) data so that they can perform well on diverse downstream datasets and tasks. The goal when training foundation models is to learn how to extract general-purpose feature representations that can be reused across different domains and applications. Similarly, the objective of foundation models in the multimodal domain is to enable learning across diverse modalities and tasks, but these models are constrained by the research emphasis on only text and image modalities. 
% Indeed, there is increasing evidence that the community will have to fundamentally re-envision multimodal
% architectures and training techniques.

\begin{table*}
	\centering
	\caption{Comparison of conventional multimodal model properties with that of generalist multimodal models}
	\label{tab:diff}
    \scalebox{0.85}{
	\begin{tabular}{|p{3.0cm}|p{6.0cm}|p{6.0cm}|}
		\toprule
		\textbf{Attributes} &
		\textbf{Typical Multimodal learning} & \textbf{Generalist Multimodal learning} \\
		\hline
		\textit{Modality} &
		Limited set of input modalities, primarily vision, text, or audio & Large set of more diverse modalities,
  including vision, language, audio, video, time series, sensor, graph, etc. \\
  		\hline \textit{Objective } &
Focus on cross modal learning & Equal importance to both unimodal and cross-modal learning \\
		\hline
		\textit{Flexibility} &
Rigid architecture in regard to encoding and decoding. & Modular architecture that allows flexible combination of modules for unimodal and multimodal tasks. \\
  \hline
  \textit{Adaptability} &
  Discrepancy between pretraining and finetuning tasks lead to low zero/few shot capability & High zero/few shot capability with being adaptive to new downstream tasks \\
  \hline
  \textit{Generalizability} &
High discriminative but low generative capability & Low discriminative and high generative capability \\
  \hline
  \textit{Popular models} &
  VATT, ALBEF, BLIP, LF-VILA, BriVL, OmniVL, LLaVA, mPLUG-owl, Kosmos-1, etc. & GATO, OFA, OFA+, MPLUG-2, META-TRANSFORMER, NEXT-GPT, etc. \\
  \hline 
	\end{tabular}
	}
\end{table*}

Motivated by these gaps, a surge of new design elements have been introduced in the multimodal literature \cite{zhu2022uni, lu2022unified, wang2022ofa, bai2022ofasys, reed2022generalist, xu2023mplug}. These are largely inspired from the success of unimodal foundation models in NLP and Vision. 
We refer to this new line of models as \emph{Generalist Multimodal models} (GMMs).
% or interchangeably \emph{foundation multimodal models (FMMs)}.
GMMs can include those that can operate across modalities beyond the two most common data types in the research, text and image. More specifically, the model must demonstrate capabilities spanning several modalities including but not limited to text, image, speech, audio, video, and sensor. This broader definition captures models with widely generalizable representations across diverse modalities.
A detailed differentiation between our definition of generalist multimodal and typical multimodal models is summarized in Table \ref{tab:diff}.
% based on the recently defined concept of a foundation model \cite{bommasani2021opportunities}.

\begin{figure*}
\centering
	\includegraphics[width=0.9\textwidth]{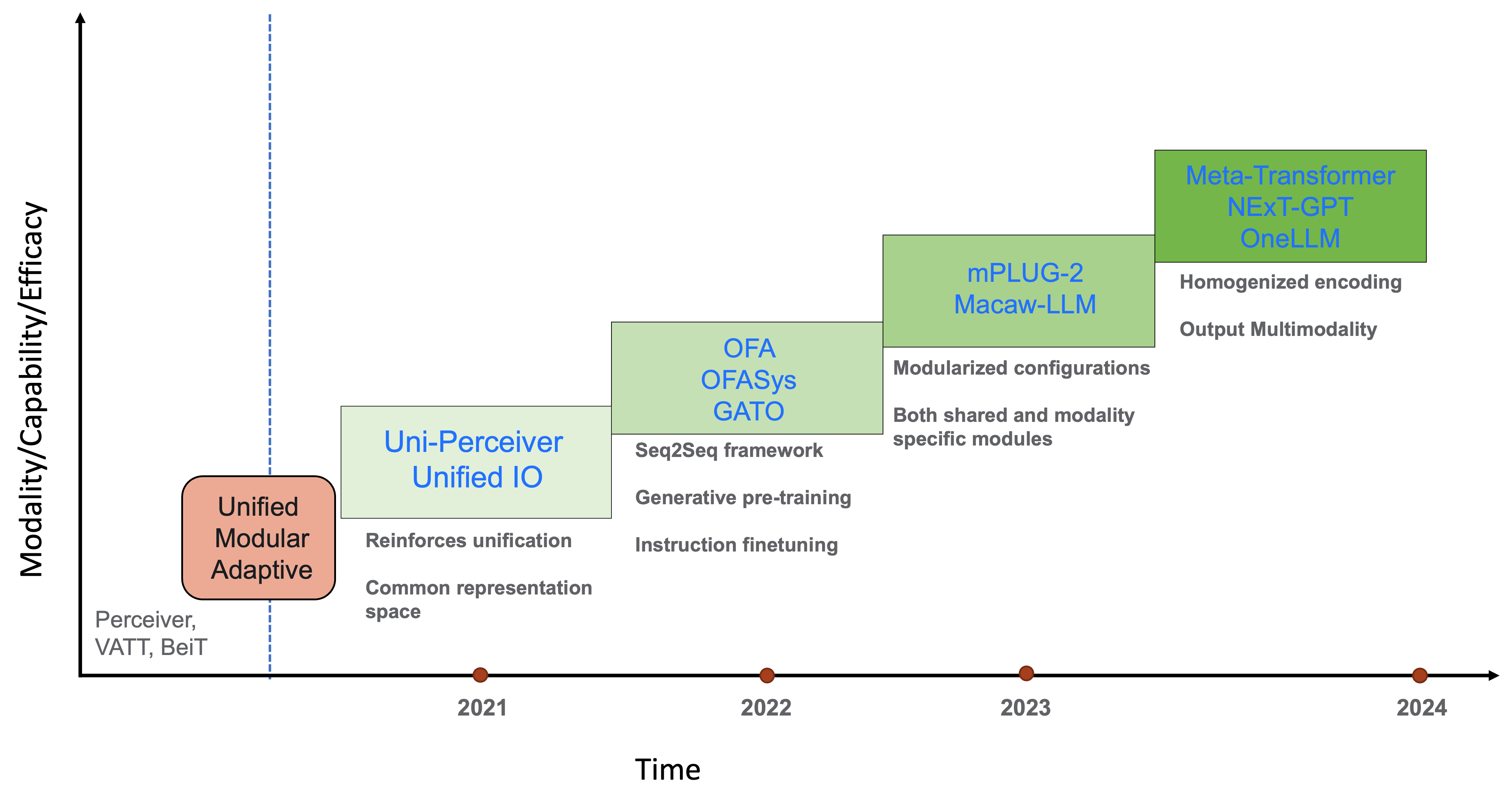}
	\caption{Evolution of GMM capability with time}
	\label{fig:EvolFMM}
\end{figure*}

In contrast to standard deep learning models, foundation models possess various unique properties, including large-scale pretraining (supervised or/and unsupervised, e.g. masked language modeling~\cite{devlin2018bert}) and special finetuning strategies (e.g., prompt tuning, parameter efficient finetuning).
% Many of these innovations are driven by large model or data size.
These properties of foundation models empower them to be front-runners in the text and vision modalities \cite{yang2023foundation}. Such characteristics are also introduced in GMMs and have shown similar improvements in multimodal learning. On the other hand, there are numerous aspects of multimodal learning with respect to architecture, training strategies, and evaluation, which make the development of GMMs a rich area of research in its own right. There is a continuous growth of capability in GMMs with novel strategies as depicted in Fig. \ref{fig:EvolFMM}. Thus, it is valuable to review current GMMs efforts and identify essential properties for further enhancing GMMs capabilities. In this review, we identify such emerging properties and holistically analyze them. 

% {\color{red}{Maybe say we are restricting ourselves to deep learning models.}}

Although there are a few survey papers on multimodal learning \cite{liang2022foundations,acosta2022multimodal,goyal2022survey,li2023multimodal}, they have the following limitations: (i) predominantly dealing with the text-vision paradigm with little to no consideration of other modalities; (ii) focusing solely on data fusion across modalities and overlooking other critical factors such as architectural designs, pretraining objectives, and an ever-growing range of multimodal tasks \cite{gao2020survey,muhammad2021comprehensive}; and (iii) a greater emphasis on crossmodal learning with little consideration to unimodal aspects \cite{liang2022foundations}.
Therefore, we provide a comprehensive review of existing GMMs (models that worked beyond text and vision) incorporating various data handling, architectural, and training aspects. To the best of the authors' knowledge, this work is the first to provide a holistic review of the recent trends in GMMs learning.
The main contributions of this review are as follows:
\begin{itemize}
\item a new taxonomy that addresses the design space of current multimodal architectures.
    \item taxonomy factors aligning explicitly with the context of foundation models, in contrast to previous survey papers.
    \item a methodology to problematize current approaches based on the taxonomy.
    \item a range of research directions that could advance the multimodal paradigm.
\end{itemize}

The rest of this paper is organized as follows: Section \ref{sect-background} provides the background on foundation models for various unimodal domains. Section \ref{sect-architecture} discusses a typical architecture pipeline of GMMs. Section \ref{sect-taxonomy} describes our taxonomy, classifies existing works into the taxonomy, and uses the taxonomy to comment on the merits and limitations of current approaches. Section \ref{sect-challenges} highlights the key challenges in the multimodal foundation paradigm, and Section \ref{sect-opportunities} lists the potential future research opportunities to enable the development of true generalist models. Finally, Section \ref{sect-conclusion} concludes the review with the summary of our findings.

\section{Background: Unimodal Foundation models}
\label{sect-background}

\subsection{Foundation Language Models (FLMs)}

The development of Foundation Language Models can be traced back to the release of BERT (Bidirectional Encoder Representations from Transformers) \cite{devlin2018bert}. BERT was a breakthrough in NLP because it was pretrained on a large corpus of text data using unsupervised training via an attention-based architecture~\cite{vaswani2017attention}, which allowed it to capture bidirectional context information and achieve state-of-the-art performance on various downstream tasks. After that, a series of models were developed to improve on BERT, each offering some unique capabilities but sharing the same attention mechanism. GPT-3 is an auto-regressive decoder only model, which shows strong performance on many NLP tasks under the few-shot and zero-shot settings \cite{brown2020language}. Similarly, T5 is an encoder-decoder model that models all text-based language tasks with a text-to-text format \cite{raffel2020exploring}. This introduces a unified framework for better context learning across tasks and serves as the foundation for several unified model architectures, such as Uni-Perceiver and Unified-IO~\cite{lu2022unified,zhu2022uni}. Recently, instruction-tuned models, which are trained to respond to human-specified instructions, have gained popularity due to their improved flexibility and interpretability \cite{chung2022scaling, sanh2021multitask, xue2020mt5}. The Flan-T5 model achieves instruction tuning via chain-of-thought prompting where a sequence of interrelated prompts  are provided in a coherent way \cite{chung2022scaling}. By fine-tuning on this type of data, Flan-T5 can learn to generate text that follows a logical flow and maintains coherence across long spans of text. Likewise, T0 has been fine-tuned using a combination of natural language descriptions and code snippets as instructions \cite{sanh2021multitask}. The model was trained on a diverse set of programming tasks and achieved state-of-the-art performance on several benchmarks. As of this writing, LLAMA~\cite{touvron2023llama} and MISTRAL~\cite{jiang2023mistral} represent open source models that offer state-of-the-art performance on various language tasks. The progressive improvement of FLMs has been driven primarily by improved training procedures, such as formulating language tasks with instructions and novel architectures, e.g. sparse attention to encode long-distance dependencies~\cite{dao2022flashattention}.

\subsection{Foundation Vision Models (FVMs)}
Convolutional neural networks (CNNs) were the state-of-the-art models for large scale vision tasks before the emergence of the vision transformer (ViT) in 2020 \cite{dosovitskiy2020image}. The ViT is based on the transformer architecture (multi head attention) ~\cite{vaswani2017attention}, which demonstrates superior performance on various computer vision benchmarks, using self-attention mechanisms to capture global dependencies between image patches. This outcome is achieved even though the ViT does not include the image priors (such as translation invariance) that are built into the architecture of CNNs. To reduce the computational cost of ViT architectures, especially for high-resolution images, \cite{liu2021swin} introduced Swin Transformer, which divides the input image into smaller patches and processes them in a multi-stage manner. In each stage, it applies self-attention mechanisms to capture global dependencies between patches and uses shifted windows to enable efficient computation and memory usage. The shifted windows allow each patch to attend only to its neighboring patches, reducing the computational cost of self-attention.
% Additionally, EVA  leverages unlabeled images with the large-scale pretrained image-text model (such as CLIP to scale up the standard ViT model.
The recently introduced model InternImage~\cite{wang2023internimage} adopts deformable convolution as the core operator rather than pure attention. This results in a model that not only has the large effective receptive field required for downstream tasks such as detection and segmentation, but also has the adaptive spatial aggregation conditioned by input and task information \cite{wang2022internimage}. Similarly, InternVideo extends the deformable mechanism to video tasks by assembling two large video models with both generative and discriminative capability \cite{wang2022internvideo}.

Additionally, there are vision-language alignment models such as CLIP~\cite{radford2021learning}, FLORENCE~\cite{yuan2021florence}, FLAVA~\cite{singh2022flava}, LLAVA~\cite{liu2023improved} and SCI-TUNE~\cite{horawalavithana2023scitune}. CLIP (Contrastive Language-Image Pre-Training) uses a contrastive pretraining objective to learn joint representations of images and text without the need for explicit alignment between the two modalities \cite{radford2021learning}. Likewise, FLAVA learns to align the semantic meaning of words and visual concepts via a novel attention mechanism, which allows the model to attend to specific regions of an image based on the meaning of the associated words. \cite{singh2022flava}. SCI-TUNE~\cite{horawalavithana2023scitune} specifically aligns vision with language space for scientific tasks. FVMs have been continuously advanced via Transformer/Diffusion architectures and cross modality alignment.

\subsection{Foundation Time Series Models (FTMs)}
Time series data from sensors represents one of the most widely used data modalities for real-life applications; it is ubiquitous, from bio-medical analysis to aerospace engineering to marketing. One  major challenge in time series analysis is a lack of model transferability between different datasets. A model produced for a specific application can rarely be reused for another unless the applications have high similarity. While  interest in developing foundation models for time series is growing, research is still mostly restricted to individual domains \cite{wu2020deep,zhou2021informer,grigsby2021long,garza2023timegpt}. The first relevant model is INFORMER~\cite{zhou2021informer}, which uses attention modules to capture long sequence dependency in time series forecasting. Specifically, it introduces a probabilistic attention mechanism to prioritize active queries instead of lazy ones and offers a sparse Transformer module to mitigate the quadratic compute and memory requirements of vanilla attention. Secondly, there is SPACETIMEFORMER~\cite{grigsby2021long}, which goes from one dimensional time domain to space-time for explicitly capturing variable relationships in multivariate time series forecasting. It formulates multivariate forecasting as a spatiotemporal sequence problem where each input token to a Transformer represents the value of one variable at a specific time. Then, Long-Range Transformers jointly learn the interactions between spatial information, temporal information, and the variable values along the extended sequence~\cite{grigsby2021long}. Third is TIMEGPT~\cite{garza2023timegpt}, a recent large scale model that demonstrates high zero shot performance for diverse time series tasks.
It is based on standard Transformer encoder-decoder architecture which takes a window of historical values to produce the forecast and adds local
positional encoding to enrich the input.  Attention-based architectures are driving FTMs
by capturing the diversity of past events and correctly extrapolating potential
future distribution.

\subsection{Foundation Graph Models (FGMs)}
As with research on extending foundation models to time series, there has been preliminary research into the development of large scale pretrained foundation models for graph-structured data, such as citation networks and protein interaction networks \cite{liu2023towards}. FGMs are pretrained on large-scale graph datasets using unsupervised or semi-supervised learning methods to learn general feature representations of graphs \cite{rong2020self,jin2021automated,das2023there}. These pretrained graph models can then be fine-tuned on downstream tasks, such as node classification, link prediction, and graph classification. By leveraging the learned feature representations from the pretrained model, the downstream task can benefit from improved performance and reduced training time. In GROVER~\cite{rong2020self}, the authors propose a pretrained graph neural network (GNN) model for predicting molecular properties. The GNN is first pretrained on a large-scale molecular graph dataset using a self-supervised learning pretext tasks, such as mask node prediction, i.e., predicting masked nodes in the graph based on their surrounding context.
After pretraining, the model is fine-tuned on tasks such as predicting solubility and toxicity of molecules. 
% They demonstrate that their pretrained GNN model outperforms previous state-of-the-art methods on several benchmark datasets.
Unlike FLMs, there are various self supervised pretext tasks in FGM, including PAIRDIS~\cite{peng2020self} and PAIRSIM~\cite{jin2021node}.
% Similarly, \cite{rong2020self} proposes a contrastive learning framework for pretraining graph neural networks (GNNs) using data augmentations. Specifically, it design four types of graph augmentations to incorporate various priors in unsupervised learning.
~\cite{jin2021automated} observed that different pretext (pretraining) tasks affect downstream tasks differently across datasets, which suggests that combining multiple pretext tasks is crucial for graph pretraining. In this regard, \cite{jin2021automated} develop an AUTO-SSL framework to automatically leverage multiple pretext tasks effectively by associating weights to tasks. Furthermore, U-SSL~\citep{das2023there} extends pretraining capability from a single graph to multiple graphs of a particular family, demonstrating advantages of multi graph learning over existing SSL works.   

\begin{figure*}
\centering
	\includegraphics[width=0.8\textwidth]{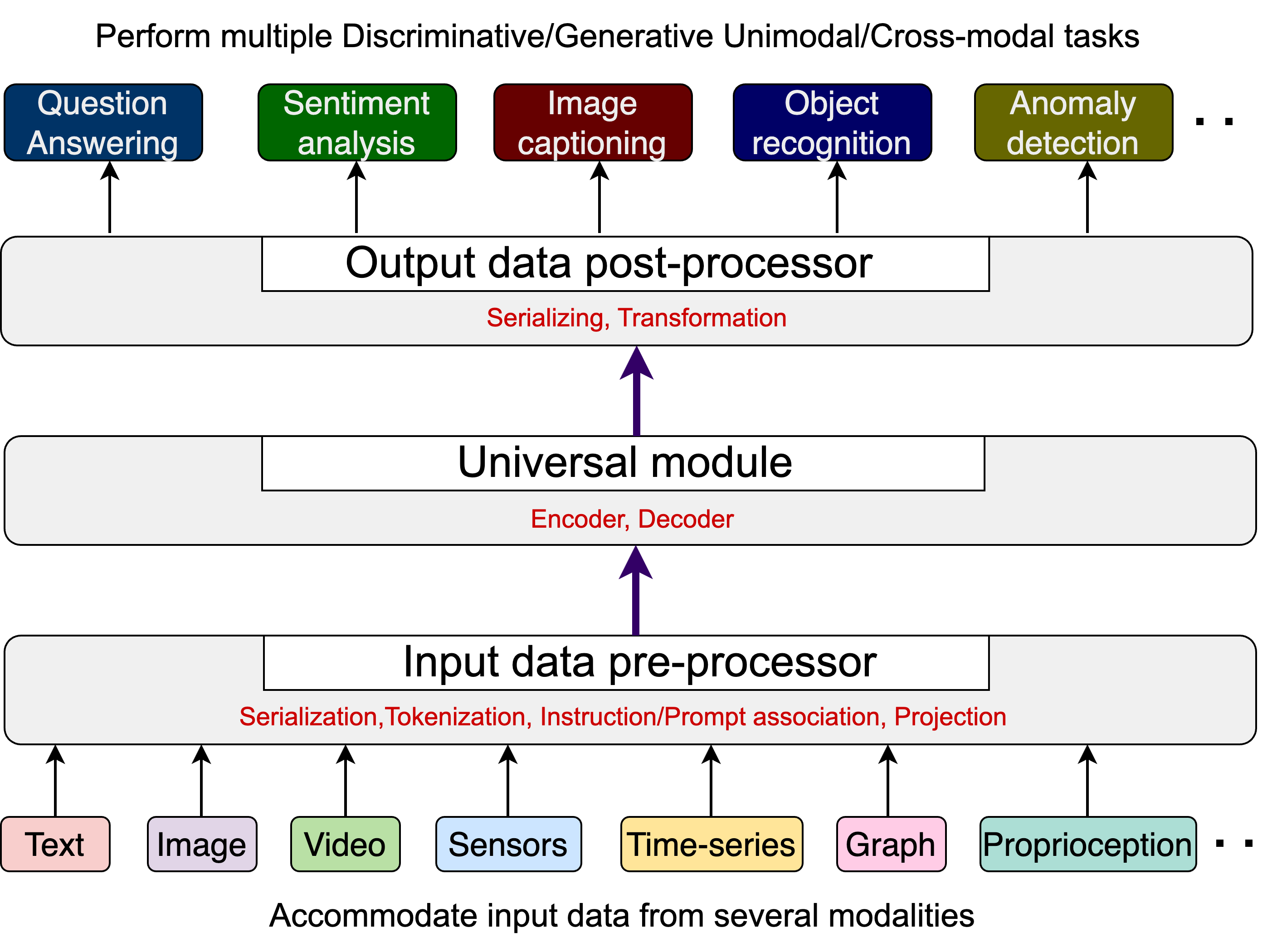}
	\caption{A generalist multimodal architecture pipeline}
	\label{fig:FMMpipeline}
\end{figure*}
\section{Generalist Multimodal model pipeline}
\label{sect-architecture}
A typical GMM architecture pipeline from input data to output predictions can be divided into different phases as described below and illustrated in Figure \ref{fig:FMMpipeline}. The following subsections describe these different phases in greater detail.

\subsection{Input pre-processing}
The first module is related to the data pre-processing where raw data from different modalities are transformed into a form that can be consumed by a universal learning model. It involves different stages which are as follows:

\subsubsection{Serialization/Tokenization} This process 
converts different modalities like text, audio, image, etc. into common numerical formats (also known as tokens). For instance, in text modality, input text is parsed into tokens where each token is mapped to a numerical ID from the model's vocabulary. For vision modality, images are resized to fixed size patches (e.g., 224 x 224 pixels in CLIP~\citep{radford2021learning}) and corresponding pixel values are stored in numerical tensors. In audio modality, raw audio is transformed into a spectrogram representation which is further chopped into small time/frequency frames. In point clouds, "farthest point sampling" (sample of a representative skeleton of original point clouds), nearest neighbor, and adjacency matrix can localize/streamline input data~\citep{zhang2023meta}. The primary objective of this step is to prepare data for the consumption by the encoder. 
% (BPE for text, grid/patches for images/video, spectrogram or MFCC for audio, etc.) 

\subsubsection{Encoding} 
Encoders fetch a numerical representation of input tokens in high-dimensional space known as embeddings. Encoders leverage the predefined knowledge (via trained frozen models) to accurately position the input token in a high-dimensional semantic space that supports learning. For text modality, any large language model (LLM) which has been trained on massive text corpora can serve as an efficient embedding model.
The family of models from CLIP and CLIP-VIT~\citep{radford2021learning} are strong candidates for encoding visual information, including images and video frames.
Large scale audio models such as WHISPER~\citep{radford2023robust} are employed for encoding audio modality. All encoders discussed above are modality-specific, which are typically trained separately, leading to potential discrepancies in the representations (embeddings) generated by different encoders. IMAGEBIND~\citep{girdhar2023imagebind} is one potential solution that learns a joint embedding across six modalities, including images, text, audio,
depth, thermal, and inertial measurement unit (IMU) data. GMMs such as NEXT-GPT leverage IMAGEBIND to encode their input modalities. Furthermore, recent GMMs, such as  META-TRANSFORMER~\citep{zhang2023meta} and ONELLM~\citep{han2023onellm}, have shown that any well pretrained transformer may serve as a universal cross-modal encoder.

\subsubsection{Projection} 
Projection transforms the representations (embeddings) from the encoders into a new space that is comprehensible by the universal model. Typically, an LLM is used as universal model; thus, the projector transforms original embeddings into language space.
While serialization, tokenization, and encodings are standardized, the projection steps vary across models and are usually trainable components. Projection could range from simple fully connected linear layers to complex convolutional operations. It also aligns different modality-specific representations via cross-attention and other elegant mechanisms.

\subsection{Universal Learning}
The unified representation from different modalities in the input pre-processing module is fed to the second module, which is a universal/backbone model that performs representation learning and reasoning in a shared semantic space through multiple neural network layers. A pretrained/fine-tuned LLM is typically used as the universal model in multimodal learning (e.g., BART in OFA~\citep{wang2022ofa}, LLAMA-2 in ONELLM~\citep{han2023onellm}). This is mainly due to two reasons: (i) unlike other modalities, language models have been extensively trained on vast amounts of data for various general purpose tasks which leads to a strong knowledge model at the center, and (ii) both input and output interactions are executed mostly in text form so it is logical to use LLMs as the central model and align other modalities around it as opposed to the inverse. 

\subsection{Output Decoding}
In the final module, the data post-processing phase transforms the learned multimodal representation into modality/task-specific outputs. The decoder utilizes the rich fusion of multimodal encoder representations to generate task specific outputs that have contextual meaning informed by cross-modality understanding in universal LLMs.
% Overall, the key idea is converting diverse inputs into a common representational space where the universal model can inductively learn alignments and patterns across modalities.
For text only output, one leverages standard Transformer decoder (with attention, cross-attention and multilayer perceptron (MLP) layers) where a shared model can take different kind of input and adapt to a variety
of tasks with text generation. Diffusion decoder models such as Stable
Diffusion (SD)~\citep{rombach2021high} for image generation and AudioLDM for audio synthesis~\citep{liu2023audioldm} are preferred.

\section{Taxonomy of Generalist Multimodal architectures}
\label{sect-taxonomy}
\begin{figure*}
\centering
	\includegraphics[width=0.95\textwidth]{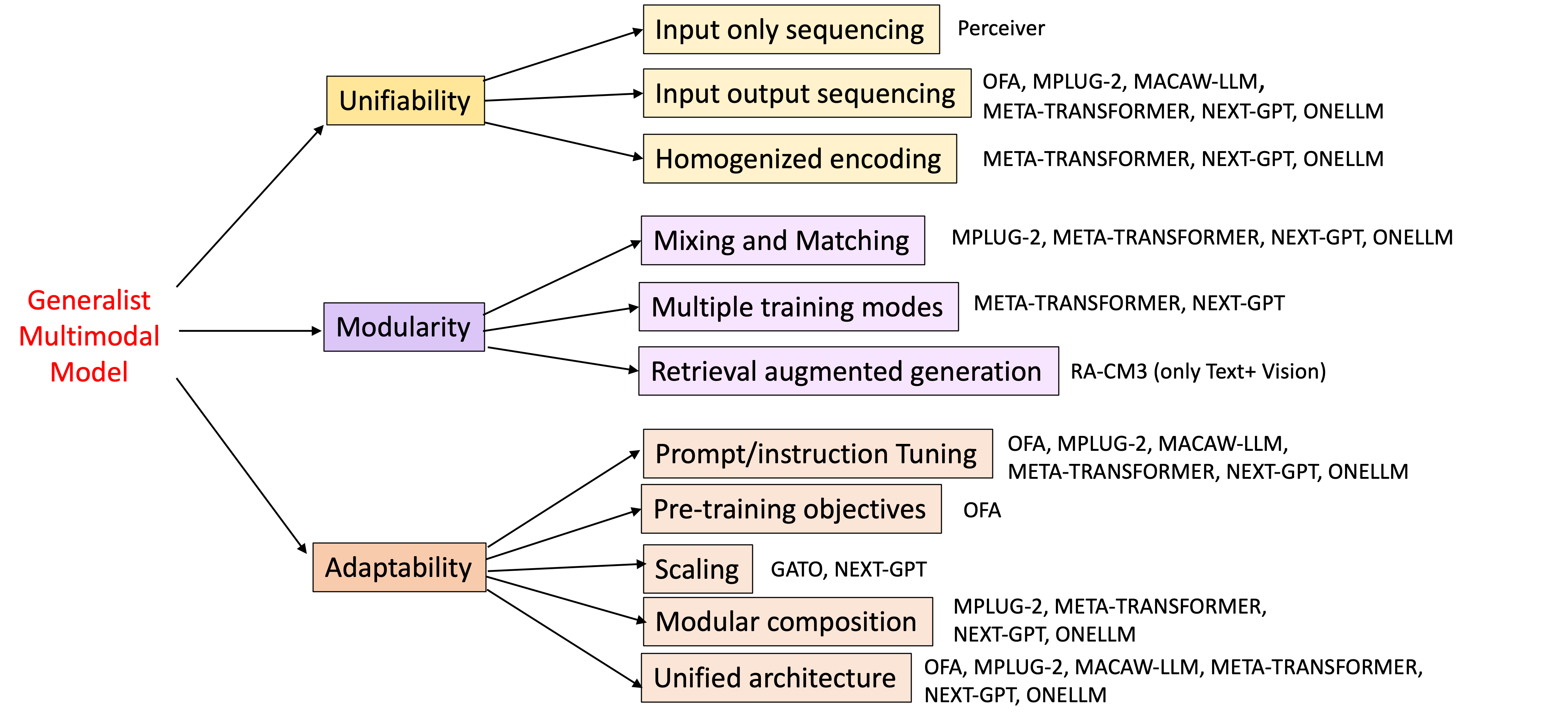}
	\caption{Taxonomy tree}
	\label{fig:TaxTree}
\end{figure*}

This section provides a novel classification taxonomy for multimodal architectures. In contrast to traditional metrics such as "Representation", "Reasoning", "Transferability", etc., we identified three novel factors for categorization: \textit{Unifiability, Modularity, Adaptability}. These factors are major driving forces of modern multimodal architectures and are relevant to the development of generalist multimodal models. The following subsections analyze existing works via the lens of these elements, and their taxonomy is illustrated in Figure \ref{fig:TaxTree}. 

\subsection{Unifiability}
\label{sec:unifiability}
Unifiability enables the encoding of the inputs and targets in a common representation space and modeling of the joint probability of inputs and targets through the similarity of their representations. Unifiability eliminates the gap due to task-specific formulations and encourages better collaborative learning between various tasks and modalities. It can be achieved via an end-to-end model architecture that converts inputs from different modalities and outputs as a sequence of discrete tokens. These tokens are drawn from a unified and finite vocabulary by leveraging techniques such as image quantization and synthesis~\citep{van2017neural,ramesh2021zero,esser2021taming}.
Thus, models can efficiently handle several modalities and tasks without any modality or task specific components. The model parameters would be shared among the modalities and tasks. Typically, unification is accomplished in three stages. The first stage is {\em tokenization}, a process that receives input from multiple modalities. The second stage comprises the backbone module, which transforms data from different modalities into a common representation space.
% It is usually a transformer-based model with several layers of transformation.
Finally, the decoder/head projects the encoded data tokens to a task-specific target space in stage three.
% {\em Provide mathematical aspects of representing/embedding different modality input in one space. How does it enables cross modal/task learning.}

The concept of unifiability originated in unimodal foundation models, such as GPT-3 \cite{brown2020language} and T5 \cite{raffel2020exploring}, due to the models' focus on learning a single representation space across multiple language tasks. Their usability was further enhanced by expanding the tasks to visual and structured inputs and outputs \cite{chen2021pix2seq,wang2021simvlm,yao2022pevl}.
The strategies for unifiability can be categorized into three phases based on their development process. In the initial phase of \textit{Input-Only Sequencing}, unification is only achieved at the modality level by framing input from different modalities as a sequence of discrete tokens. Thus, a model would be modality agnostic; although, there will be task-specific modules (heads). In the second phase of \textit{Input-Output Sequencing}, unification is further improved by generalizing it across modalities as well as tasks. This is achieved by representing inputs and outputs as a sequence of tokens. Recently, a third phase of \textit{Homogenized encoding} was introduced, where the unification is further strengthened by
utilizing a shared encoder/projector across modalities.
% associating task-based instructions with the input sequences. These strategies are discussed in detail in the forthcoming subsections.

\subsubsection{Input-Only sequencing}
In this strategy, different modalities are handled via serialized input tokens followed by a universal model and decoders. Different modalities are tokenized (serialized) and then projected into language space. 
PERCEIVER (and its variants) belongs to this class \cite{jaegle2021perceiver, jaegle2021perceiverio}.  It processes information from multiple modalities as a uniform byte-sequence representation which becomes typically large for some modalities such as ImageNet images, which contain $50176$ pixels (i.e., input data points) at $224\times 224$ resolution.
% Since sequencing all input tokens losses the spatial/position information, it is explicitly incorporated by tagging position encodings with input tokens.
Therefore, PERCEIVER~\cite{jaegle2021perceiver} proposes a cross-attention module to encode the input array using a smaller number of latent feature vectors. Thus, along with modulating the dimension of the input vector, which is usually carried out in Transformers~\cite{vaswani2017attention}, it also reduces the number of latent vectors corresponding to the input. This reduces the risk of the runtime and memory explosion that can happen with a large number of input tokens, and it allows the model to apply attention to very large and generic inputs while using a deeper architecture. However, this introduces an attention bottleneck and may restrict the model's ability to capture all of the essential details from the input signal. Therefore, Perceiver applies cross-attention and the Transformer operation iteratively, which allows the latent array to extract information from the input as needed. Finally, it decodes by applying an attention module to map from latent arrays to output arrays. The primary contribution of this work is in terms of handling different modalities via sequencing and enabling better contextual learning via direct application of attention modules on the input. However, this strategy for unifiability is not  effective for handling multiple tasks off-the-shelf  and significantly falls behind the baselines in generative tasks.   

\subsubsection{Input-Output sequencing}
The true unifiability seems to be achieved in this stage where 
along with inputs, different task targets are  serialized into unified token sequences. UNIPERCEIVER~\cite{zhu2022uni} is a preliminary work in this class where input is first serialized with modality-specific tokenizers and then encoded via modality-specific encoders into the shared representation space. Any task is modeled as finding the maximum likelihood target for each input through the similarity of their representations. The main contribution lies in representing all modalities and tasks in a common representation space, enabling high zero-shot capability. However, its decoding mechanism is not elegant, as it uses a PERCEIVER decoder based on cross-attention where decoding is performed implicitly via queries. As a result, its performance suffers in downstream tasks that are different from pretraining, and it exhibits poor generative capability. To overcome this shortcoming, UNIFIED-IO \cite{lu2022unified} introduces a specific decoder, where projected multimodal embeddings are appended to the input sequence to assist the decoder in predictions. Similar Input-Output sequencing mechanisms can be found in other GMMs such as MPLUG-2, OFA, MACAW-LLM, META-TRANSFORMER, etc. Of note, most GMMs before META-TRANSFORMER~\cite{zhang2023meta}, leveraged modality specific encoders for representing modality data. However, META-TRANSFORMER~\cite{zhang2023meta} is the first framework that employs
a unified encoder to simultaneously extract representations from multiple modalities with the same set of parameters. Specifically, META-TRANSFORMER first transforms multimodal data into a sequence of tokens and then a modality-shared encoder with frozen parameters extracts representations, which are further adapted to individual tasks by updating the
parameters of downstream task heads and lightweight tokenizers only. The mapping of modalities into shared token a space removes the need for paired training data (such as image-text pairs), which endows multimodal
learning with more training flexibility. Unlike META-TRANSFORMER, NEXT-GPT~\cite{wu2023next} leverages a single multimodal encoder ImageBind~\cite{girdhar2023imagebind} to encode all modalities, i.e., text, image, audio and video, into a language space rather than leveraging modality-specific encoders.

The predominant strategy for unifiability suggests sharing the same encoding, backbone, and decoding modules across all modalities. However, this strategy may suffer from the issue of modality entanglement when expanded to several modalities (beyond text and images) due to the large variance of different modality tasks~\cite{xu2023mplug}. The challenge is that multiple modalities may interfere with each other, especially when there are numerous modalities and tasks~\cite{huang2022modality}. 
MPLUG-2~\cite{xu2023mplug} takes a hybrid unifying approach that
can balance the gain of modality collaboration and the influence of modality entanglement. It has both  shared functional modules to encourage modality collaboration and modality-specific modules to address the problem of modality entanglement. Specifically, it designs a unified text encoder and dual-vision encoder module by disentangling spatial and temporal representations. Video inputs share the Transformer module with image inputs for modeling spatial information and an additional local temporal module for temporal relation modeling on video-related tasks. Then, a novel universal layer module is introduced to serve as a pivot across different modalities, where vision and language modalities are projected to the common language-guided semantic space by sharing self-attention modules. Finally, it has a multimodal fusion module and a shared decoder module for unimodal and crossmodal generation, respectively. It has a unified pretraining objective comprised of language loss such as mask language modeling \cite{devlin2018bert}, cross-modal matching loss \cite{li2021align}, and instruction-based language model loss \cite{alayrac2022flamingo,wang2022ofa}. MACAW-LLM~\cite{lyu2023macaw} proposes a novel unification approach where input multimodal features are directly aligned to the embeddings of a universal LLM in a one-stage instruction fine-tuning process, as opposed to all other GMMs which align and fine-tune in two separate training process. 

% Similar to Uni-Perceiver, it is a sequence-to-sequence unified architecture that performs large and diverse tasks in a modality/task-agnostic manner. 
% The unified architecture does not need specific task or modality branches, which
% is accomplished by homogenizing the input and output of each task into a sequence of discrete vocabulary
% tokens.
% It trains a single transformer-based architecture on over 90 diverse datasets in the vision and language
% fields. UNIFIED-IO is the first model to perform various tasks and produce strong results across 16 diverse
% benchmarks without finetuning. Unified-IO \cite{} follows this design but applies it to a wider set of tasks than previous works, including keypoint estimation, image in-painting, and
% region captioning. 
% \subsubsection{Task-fused sequencing}
Task-fused sequencing is a more advanced mechanism where task-relevant information, such as an instruction, is associated with the input sequence for better representation learning.
OFA~\cite{wang2022ofa} is one of the first GMMs that introduces the
% Similar to UNIFIED-IO and UNI-PERCEIVER, OFA \cite{wang2022ofa} is also a sequence-to-sequence unified learning framework for pretraining, finetuning, and inference on all tasks concerning different modalities. However, we mark it as a third-generation unification strategy since it associates task-relevant instructions with the input sequence for better representative learning. Transformer with an encoder decoder acts as the backbone architecture for all the pretraining, finetuning, and zero-shot tasks. Both pretraining and downstream tasks of cross-modal and uni-modal understanding/generation are all formed as Seq2Seq generation. 
% It is available to perform multitask
% pretraining on multimodal and uni-modal data to endow the model with comprehensive capabilities.
% The unique feature of OFA is its 
instruction-based task specification that leads to superior performance in various unimodal and multimodal tasks. It formulates both pretraining tasks and downstream tasks of crossmodal
and unimodal understanding and generation as Seq2Seq generation.
Specifically, it shares the identical schema across all tasks, while specifying handcrafted instructions for discrimination. For instance, in one of the visual grounding tasks, the model
learns to generate location tokens specifying the region position $x1; y1; x2; y2$ based on the input of the image $x^{i}$ and
the instruction “Which region does the $x^{t}$ describe?” where $x^{t}$ refers to the region caption. Similarly for captioning task, the model learns to generate a description based on the input image $x^{i}$ and the instruction “What does the region describe? region: $x1; y1; x2; y2$. This instruction-based training unifies multiple modalities and multiple tasks into a single model and pretraining paradigm. This mechanism enables a single model to perform different task, or even new ones specified via instructions, across modalities.

Another elegant task-fused sequencing mechanism can be seen in NEXT-GPT~\cite{wu2023next}, whose universal model (i.e., Vicuna~\cite{chiang2023vicuna}) provides signals tokens corresponding to each modality along with the text generation. These signal tokens serve as instructions to activate corresponding decoding layers and dictate what content to produce if activated. In this regard, various downstream multimodal decoders are being used, such as Stable
Diffusion (SD) for image synthesis~\cite{rombach2021high}, Zeroscope for video synthesis~\cite{zeroscope}, and AudioLDM for audio synthesis~\cite{liu2023audioldm}. The signal representations are fed into the condition encoders of the conditioned diffusion models for generation. In a nutshell, NEXT-GPT enables unification across both input and output modalities, capable of generating text, image, audio and video, which is not seen in any previous GMMs. 

\begin{table*}
\caption{GMMs that implement unifiability strategy}
\label{Tab:unif-strat}
\begin{tabular}{|c|c|}
\hline
\textbf{Strategy}                                                           & \textbf{Models}                                                                                     \\ \hline
\textbf{\begin{tabular}[c]{@{}c@{}}Input-Only\\ sequencing\end{tabular}}    & \begin{tabular}[c]{@{}c@{}}Perceiver, Perceiver-IO \\  Hierarchical Perceiver\end{tabular}          \\ \hline
\textbf{\begin{tabular}[c]{@{}c@{}}Input-Output \\ sequencing\end{tabular}} & \begin{tabular}[c]{@{}c@{}}Uni-Perceiver, Unified-IO\\ META-TRANSFORMER\\ MPLUG-2, OFA\end{tabular} \\ \hline
\textbf{\begin{tabular}[c]{@{}c@{}}Homegenized\\  encoding\end{tabular}}    & \begin{tabular}[c]{@{}c@{}}Meta-Transformer, \\ NExT-GPT, OneLLM \end{tabular}                       \\ \hline
\end{tabular}
\end{table*}

\subsubsection{Homogenized encoding}
Most GMMs leverage frozen modality-specific encoders for obtaining modality embeddings. Due to various training scopes, these embeddings lie in different representation spaces, which degrades the alignment process in the projector and universal modules. Therefore, it is desirable to use encoders that can represent several modalities in one space. META-TRANSFORMER~\citep{zhang2023meta} is the first GMM to demonstrate that a frozen encoder (visual model ViT~\citep{dosovitskiy2020image} in this case) can achieve competitive results across several ($12$)
data modalities, including text, images, point clouds, and audio. This is made possible due to modality-specific data to sequence trainable tokenizer that projects multimodal data into token sequences that share a common manifold space. The learning representation from the encoders is adapted to individual tasks by updating the parameters of the task-specific heads. Similarly, ONELLM~\citep{han2023onellm} leverages light weight modality-specific tokenizers (each comprising of one 1D/2D convolutional layers) to convert the input signal into a sequence of tokens and then employ the frozen pretrained model (CLIP-ViT~\citep{radford2021learning}) as universal encoder for all modalities. 
% Particularly, it uses CLIP-ViT~\citep{radford2021learning} as encoder. 
Along with universal encoder, it also introduces the universal projection module (UPM) that consists of multiple image projection experts as a universal X-to-text interface. Furthermore, it designs a modality router to control the weight of each expert for the given inputs, which
turns the UPM into a soft mixtures-of-experts. 
For the alignment stage, they train the modality tokenizers and UPM while keeping the LLM frozen. For the instruction tuning stage, they
train only the LLM and keep other modules frozen.

In short, unifiability is a core desirable feature, and it is continuously enhanced through strategies such as sequence to sequence architecture and universal encoder, among others. These strategies are summarized in Table \ref{Tab:unif-strat}. Unifiability benefits GMMs by encouraging collaboration between modalities/tasks and improving adaptability for new tasks.

\subsection{Modularity}
\label{sec:modularity}
Modularity refers to the principle of building  multicomponent model architectures designed for diverse functionality. Each module is designed to perform a specific function or task and can be combined with other modules in several ways to create different configurations or functionalities.
This enables flexibility in leveraging trained models to achieve satisfactory downstream performance in a computationally efficient manner. For instance, modularity can be seen as a core design element in recent vision-language multimodal models where state-of-the-art performance is achieved by aligning frozen vision encoders with trained LLMs via training adapter modules~\cite{liu2023visual,liu2023improved,ye2023mplug}.
% The modular components in a typical GMM constitutes: (i) modality specific tokenizers or/and encoders, e.g., Byte pair encoder for text and Resnet/CLIP-based encoder for vision; (ii) generative vs discriminative task-specific decoders; and (iii) universal backbone (LLAMA, MISTRAL) for common representation space.
The strategies for incorporating modularity can be broadly divided into three categories as described below: \textit{mixing and matching modules}, \textit{multiple training configurations}, and \textit{retrieval augmented generation}.

\subsubsection{Mixing and matching modules}
In this approach, modularity is ensured by composing model architectures with different type of components (encoders/universal model/decoders) in a plug-and-play fashion. For instance, separate encoders for each modality like text, image, audio etc., allow mixing different encoder types. Further, each component can be replaced depending upon the task and modality without impacting the entire model. Additionally, individual modules can be swapped independently as new, advanced versions become available, instead of having to retrain the entire model (e.g., replacing LLAMA with LLAMA-2 as the universal model). Furthermore, rather than just consuming output from the last module of the model, output can also be extracted from different positions of the architecture pipeline depending on the task.
This idea is primarily introduced by MPLUG-2~\cite{xu2023mplug}, which employs shared universal modules (e.g., LLMs) that encourage modality/task collaboration and modality-specific components to tackle the problem of modality/task entanglement.
% Secondly, modularity can allow for more efficient training and better performance on specific tasks by leveraging different compositions of modules for both unimodal and cross-modal tasks.
% Modularity is achieved by sharing common universal modules and 
% disentangling modality-specific ones
Specifically, MPLUG-2 has a separate text and dual-vision encoder for processing language and image/video modality, respectively. Then, a universal layer module is introduced to serve as a pivot across
different modalities, where vision and language modalities are projected to the common language-guided semantic space by sharing self-attention modules. For unimodal text-based tasks, the output is extracted from universal module. Whereas for cross-modal tasks, the output is consumed from the additional cross-modality fusion module. Similarly, in NEXT-GPT~\cite{wu2023next}, several pretrained components comprising of frozen encoder (IMAGEBIND~\cite{girdhar2023imagebind}), language model (VICUNA~\cite{chiang2023vicuna}) and decoders (Zeroscope, AudioLM) are combined via a trainable adapter module in a unified pipeline.

Furthermore, when new modalities emerge, specialized modules can be developed and seamlessly integrated into the model architecture without having to redesign the whole system. This can be seen in META-TRANSFORMER~\cite{zhang2023meta} where a new modality can be represented in a common manifold space via their novel meta-tokenization scheme and unified encoder.

% Furthermore, modules like modality encoders and projectors allow modality/task-specific customization without retraining the whole model. One can update frozen modality encoders as per the use case without disturbing the entire model. 
% The flexibility to use the outputs from different components induce enormous adaptability to uni-modal/cross-modal tasks. 
% The output of universal layers is applied to conduct unimodal
% discrimination tasks. For cross-modal tasks, an additional
% fusion module will be applied to produce cross-modal
% representations. 

\subsubsection{Multiple training configurations}
Along with combining several universal and modality-specific components, modularity can also be seen in the training process. The components can be used in frozen or/and training modes in different stages depending upon the goal of training.
MPLUG-OWL~\cite{ye2023mplug} and MPLUG-DOCOWL~\cite{ye2023mplugdoc} propose a modularized training pipeline that enhances the multimodal abilities of vision-language models. The training paradigm of MPLUG-OWL adopts a two-stage method for aligning image and text, which learns visual knowledge with the help of LLMs, while maintaining and even improving the generation abilities of LLMs. In the first stage, the visual knowledge and abstractor modules are trained with a frozen LLM module to align the image and text. In the second stage, language-only and multimodal datasets are used to jointly fine-tune a low-rank adaption (LoRA) module on the LLM and the abstractor module by freezing the visual knowledge module. In GMMs space,  META-TRANSFORMER~\cite{zhang2023meta} first pretrains the different components (i.e., vision encoder and language model) and then keeps them frozen while training the modality-agnostic encoder and fine-tuning the universal LLM model. Similar training techniques are observed in NEXT-GPT~\cite{wu2023next}.

% Finally, the uni-modal and cross-modal
% representations can be incorporated as input to a shared
% Transformer decoder for various generation tasks.

\subsubsection{Multimodal retrieval augmentated generation}
Retrieval augmented models (RAG) consist of two modular components, i.e., Retriever and Language model. The retriever model retrieves relevant context (chunks/passages/documents) from a corpus of training data by encoding the data into vector form and computing the data's similarity with the embeddings of the training data. The context is provided along with the query to the language model. This mechanism distributes the learning from the parameters space to external data storage, which allows rapid adaptation to new data just by updating the vector database (index new data) but without complete fine-tuning. Furthermore, RAG grounds the model and improves few-shot learning on new tasks. RAG enables modularity by augmenting the existing model capability with any external/new knowledge sources. It is also modular in terms of mixing/matching Retriever and Language model in various configurations such as: (i) both can be off-the-shelf models in the frozen state, (ii) both are trainable, and (iii) one would be trainable and the other would be frozen. This modularity enable us to develop  models in a computationally efficient and adaptable way. RAG with various configurations is extensively explored in language models \cite{izacard2022few,khattab2022demonstrate,hofstatter2023fid,munikoti2023atlantic}. There is a growing interest in extending RAG to vision-language models such as RA-CM3~\cite{yasunaga2023retrieval} and REVEAL~\cite{hu2023reveal}. However, no such work exists in the GMMs space. RA-CM3 leverages the text and image encoder of CLIP~\cite{radford2021learning} to retrieve relevant vision-text documents from the multimodal corpus and then feed them along with the query to CM3~\cite{aghajanyan2022cm3} generator. The results demonstrate that RA-CM3 significantly
outperforms baseline vision-language models such as DALL-E~\cite{ramesh2021zero} and CM3 on both image and caption generation tasks.

In summary, modularity is a crucial element for GMMs, and offers several advantages. First, it allows for greater flexibility and scalability in designing models. Individual modules can be developed and tested independently before being integrated into the larger model, making it easier to maintain and update the model over time. Moreover, individual models trained for different tasks can be integrated together to accomplish new tasks. Second, it enables adaptability to a new modality/task without requiring major changes in model architecture. Implications related to architecture adaptability are discussed in the next section \ref{sec:adaptability}.
Finally, modularity enables representation and interactions to be analyzed in an independent manner leading to better interpretability. For instance, one can reason out the GMMs performance in terms of encoder and universal model by identifying the contribution of each components. The modularization strategies adopted in the existing GMMs are listed in Table \ref{Tab:modul-strat}. 

\begin{table*}
\caption{GMMs that implement modularization strategy}
\label{Tab:modul-strat}
\begin{tabular}{|c|c|}
\hline
\textbf{Strategy}                                                                    & \textbf{Models}                                                                        \\ \hline
\begin{tabular}[c]{@{}c@{}}Mixing and matching\\ modules\end{tabular}                & \begin{tabular}[c]{@{}c@{}}MPLUG-2, META-TRANSFORMER, \\ NEXT-GPT, ONELLM\end{tabular} \\ \hline
\begin{tabular}[c]{@{}c@{}}Multiple training \\ configurations\end{tabular}          & META-TRANSFORMER, NEXT-GPT                                                             \\ \hline
\begin{tabular}[c]{@{}c@{}}Multimodal retrieval \\ augmented generation\end{tabular} & RA-CM3 (only Text+Vision)                                                              \\ \hline
\end{tabular}
\end{table*}

\subsection{Adaptability}
\label{sec:adaptability}
As data modalities and emerging capabilities of GMMs increase, so, too, does the scale of tasks. Furthermore, its expensive to retrain large models with billions of parameters for specific tasks and modalities. Therefore, models must be readily adaptable to new tasks and modalities. 
% Adaptability is also referred to as transferability.
% Adaptability can be induced in several ways, and it is one of the trending research topics in the ML community. 
Broadly, there are two ways of inducing adaptability in GMMs. It is either via training schemes (such as \textit{pretraining objectives}, \textit{instruction/prompt tuning}, and \textit{scaling}) or architectural advancements (such as \textit{unifiability} and \textit{modularity}).
We describe each of these adaptability strategies in the following paragraphs.

% \subsubsection{Few shot learning}
% Few shot learning (FSL) is a process that enables a pre-trained model to generalize over new tasks or the distribution of data (with limited samples) that the pre-trained model has not seen during training, using only a few new samples. It is a special case of inductive transfer learning, which aims to leverage knowledge learned from a distribution of data/task to improve performance on another related data distribution/task. 
% It enables adaptability to new tasks by allowing the model to fine-tune its parameters (all or some portion) on a limited number of new demonstrations. 

% The main components of FSL are task description and new demonstration (examples). These components assists model to quickly learn the new task and consequently provide predictions for new test cases. Few shot learning is often work in conjunction with other adaptation techniques.

\subsubsection{ Effective training objectives}
The choice of training objectives can have a significant impact on the adaptability of the resulting model. Objectives tailored for a narrow set of tasks may lead to overspecialization and poor generalization. Therefore, having broader objectives during pretraining is a best practice. 
A common pretraining objective of masked language modeling is randomly masking some percentage of text tokens (e.g. $15\%$) and training the model to predict the original tokens based on context. It
encourages the model to learn more robust representations that can handle missing or incomplete information. Along with standard language modeling loss, GMMs also involve cross modality objectives such as cross-modal matching loss~\citep{li2021align}, which aligns vision to language via contrastive loss and "Momentum Distillation (MoD)" to leverage a larger uncurated web dataset. During training, ~\citep{li2021align} keeps a momentum version of the model by taking the moving-average of its parameters and employing the
momentum model to generate pseudo-targets as additional supervision. With MoD, the model is not penalized for generating other reasonable outputs that are different from the annotated ground truth. Similarly, OFA~\citep{wang2022ofa} introduces instruction-based multimodal loss functions to discriminate task and modalities. In particular, they use handcrafted
instructions for discrimination, which
include video/image-text pairs, video/image captioning,
video/image question answering, text generation, etc.
% This can be useful for tasks where some inputs may be missing or noisy, such as speech recognition or natural language understanding. 
% For example, pretraining on a language modeling objective teaches the model to predict the next word in a sequence given the previous words. This task requires the model to learn contextual relationships between words and phrases, which can be useful for a wide range of downstream tasks such as sentiment analysis or machine translation.
% Similarly,
%  In most of the GMM, LLMs form an important component (e.g. the ``cognitive module'' in \cite{lyu2023macaw}), choosing the right pretraining objective can improve the model's overall ability to process multimodal data that is associated with unexpected or complex text prompts.
Furthermore, task comprehension, i.e., enough task variability for various unimodal and multimodal tasks, during pretraining ensures robust generalization \cite{wang2022ofa}. Moreover, curriculum learning (gradually increasing the difficulty of the tasks over time) helps ensure that the model learns more general features before moving on to more complex tasks~\cite{reed2022generalist}.

% Employing adhoc techniques such as regularization and dropout. Regularization can be used during both pretraining and finetuning to prevent overfitting and improve generalization performance. For instance, models such as Uni-Perceiver, Unified-IO, GATO and mPLUG-2 involve dropout regularization in their training to prevent the model from relying too heavily on any one feature or input.

\subsubsection{Instruction tuning} Instruction tuning is the process of training models to follow a set of instructions or rules that are expressed through text. In this approach, the model is provided with a set of instructions during fine-tuning that guide its behavior and decision-making process. These instructions can be handcrafted by domain experts or generated automatically using techniques like reinforcement learning. The goal of instruction tuning is to improve the model's adaptability to new tasks without relying on explicit examples or labeled data. OFA~\cite{wang2022ofa} is a classic example where instruction-based learning is used for enhancing adaptability. During pretraining, the model is trained to follow a set of handcrafted instructions for different tasks and modalities, including image classification, visual grounding, and image captioning. To adapt to a new task such as object detection in images, the developers first formulate a set of handcrafted instructions for the object detection task. These instructions might include commands like "identify the location of objects in the image" or "output bounding boxes around each detected object." These instructions are used to fine-tune the pretrained OFA model by providing the model with input images and ground-truth labels for each detected object.
% The model is trained to follow the handcrafted instructions and output predicted bounding boxes and class labels for each detected object.
This is how OFA quickly adapts to new tasks without requiring any task-specific layers or modifications to the underlying architecture. Similarly, MPLUG-2~\cite{li2022mplug} leverages handcrafted instructions to adapt to new tasks and modalities but with a more comprehensive instructions set than OFA~\cite{wang2022ofa}, spanning video/image-text Pairs, video/image captioning,
video/image question answering, text generation, etc. NEXT-GPT~\cite{wu2023next} introduces a new instruction tuning approach for cross-modal generation known as "modality switching instruction tuning." This approach equips the system with sophisticated cross-modal semantic understanding and
content generation. Specifically, an LLM takes as input the representations from different
modalities and then outputs text tokens along with modality signal tokens. Signal tokens serve as instructions to dictate which decoder needs to be activated and what content to generate. These signal tokens are passed to the corresponding diffusion decoders for content generation.
To address the absence of such cross-modal instruction tuning data, NEXT-GPT~\cite{wu2023next} manually collects and annotates a "Mosit" dataset.

\subsubsection{Prompt tuning}
Prompt tuning is a method that involves inserting specially designed natural language tokens, called prompts, into the input sequence as hints for target tasks. These prompt inputs are used to query a model and methods have been proposed to automate the prompt engineering process. These prompts can be either hard or learnable (``soft''). Hard prompts are fixed and cannot be updated during training, while learnable prompts are updated through gradient back-propagation. Prompt tuning can be much better than traditional fine-tuning in few-shot scenarios. Though instruction tuning and prompt tuning are two related techniques for fine-tuning models, there are fundamental differences between the techniques. Prompt tuning involves generating prompts or templates that are used as input to elicit specific types of responses from the model, whereas instruction tuning involves natural language instructions as input to the model to perform better on unseen tasks. A relevant example of prompt tuning can be seen in UNI-PERCEIVER~\cite{zhu2022uni} for the few-shot image classification task. The authors use a small amount of additional data (around $1\%$ of the original mini ImageNet training dataset from each class, i.e., around six samples per class out of $100$ classes) and prompt tuning to fine-tune the model. Specifically, they insert a prompt token into each input sequence that indicates the class label of the corresponding image. For example, if an input image belongs to the "horse" class, they insert a prompt token "$<horse>$" before the image tokens in the input sequence. Then, they optimize a small number of additional parameters (learnable heads) for each task. The results demonstrate that prompt tuning significantly improves performance on few-shot learning tasks (improves accuracy from $33.68\%$ to $44.12\%$) compared to pretraining alone. 
% Thus, prompt tuning allows models to quickly learn new tasks and adapt to different scenarios without requiring plenty of training data or significant modifications in architecture.

\subsubsection{Scaling}
Scaling up the model's capacity is an implicit approach to improve adaptability since it enables the model to use representations learned from diverse training data at test time, which leads to better adaptability with fewer samples. Larger models have more parameters and can learn more complex representations of the data, which can be useful for generalizing to new tasks as evident in various GMMs scaling experiments \cite{reed2022generalist,lu2023empirical,liu2023improved, wang2022ofa}. GATO~\cite{reed2022generalist} demonstrates the adaptability via scaling of data, compute, and model parameters. The model is trained on the widest variety of relevant data possible, including diverse modalities such as images, text, proprioception, joint torques, button presses, and other discrete and continuous observations and actions. GATO~\cite{reed2022generalist} is a single, large transformer sequence to sequence model with around $1.2$ billion parameters. In the RGB stacking benchmark task, GATO is initially trained on $500$ demonstrations of stacking a blue object on a green object using a 3D mouse on a real robot arm. However, this task only tested GATO's ability to adapt to perceptual variations and permutations in the objective specification. To further evaluate GATO's adaptability, authors augmented the training data with additional examples that included different colors and shapes of objects (cylinders instead of cubes). The results demonstrate that it was able to perform significantly well on new tasks compared to the previous case and that having more model parameters enhanced the model's ability to adapt to augmented training data. Scaling is not only related to model parameters but also to data sizes. For instance, \cite{lu2023empirical} observed that the higher resolution of images in multimodal learning
consistently yields $2$ to $3$ points improvement across all LLM sizes. Similarly, \cite{liu2023improved} improves \cite{li2023multimodal} by employing visual encoder (CLIP) trained on relatively larger resolution images.

Additionally, there is \textit{Low rank Adaptation} (LoRA)~\citep{hu2021lora}, which enables one to work with very large scale models in a computationally efficient manner. LoRA freezes the pre-trained model weights and injects their smaller rank decomposition matrices into the layers of the model architecture. During fine-tuning, only the smaller weight matrices are fine-tuned which greatly reduces the number of trainable parameters for downstream tasks. While it is widely adopted in FLM space, it has not been fully explored in GMMs. NEXT-GPT~\citep{wu2023next} is the only work that leverages LoRA to fine-tune the  projection layers and certain LLM parameters during instruction tuning phase. 

\subsubsection{Unified architecture} In addition to providing an end-to-end framework across modalities and tasks (described in \ref{sec:unifiability}), unified model architecture also improves adaptability in a substantial manner. 
Architecture unification by homogenizing every input and output into a sequence of discrete vocabulary tokens enables training and inference across tasks and modalities. The model is trained to map inputs and outputs from different tasks and modalities into a common sequence of discrete vocabulary tokens. During inference, the model can then use this shared vocabulary to generate outputs for new tasks or modalities, even if it has not seen examples of those specific inputs and outputs during training.
Unification through task/modality agnostic modules serve as a simple yet effective approach for adaptability.
% UNIFIED-IO~\cite{lu2022unified} is trained on referring expression comprehension tasks (text to image) and then tested on natural language inference tasks (pure text). It achieves competitive performance on the NLI tasks despite not being explicitly trained on them. 
% The higher performance was achieved because of their common sequential representation for images (images and segmentation masks are represented as sequences using a vector quantization variational auto-encoder (VQ-VAE), sparse structured outputs such as bounding boxes and human joint locations are transcribed into sequences of coordinate tokens) and text (language outputs are converted to sequences using byte-pair encoding). 
For instance, the unified architecture of OFA~\cite{wang2022ofa} enables a transfer to unseen tasks with new task instructions. The authors designed a new task called grounded question answering, where the model is given a question about a certain region on an image and must provide a correct answer. Despite not being explicitly trained on this task, OFA was able to achieve satisfactory performance. Additionally, OFA~\cite{wang2022ofa} was able to solve tasks with out-of-domain input data, such as achieving satisfactory performance in VQA for out-of-domain images without fine-tuning. Similar capabilities can be observed in other unified GMMs. 

\begin{table*}
\caption{GMMs that implement adaptation strategy}
\label{Tab:adap-strat}
\begin{tabular}{|c|c|}
\hline
\textbf{Strategy}               & \textbf{Models}                                                                                              \\ \hline
\textbf{Prompt-Tuning}          & UNI-PERCEIVER                                                                                                \\ \hline
\textbf{Pretraining objectives} & OFA                                                                                                          \\ \hline
\textbf{Instruction-Tuning}     & \begin{tabular}[c]{@{}c@{}}OFA, MPLUG-2 , \\ MACAW-LLM, NEXT-GPT,  \\ META-TRANSFORMER,  ONELLM\end{tabular} \\ \hline
\textbf{Scaling}                & GATO, NEXT-GPT                                                                                               \\ \hline
\end{tabular}
\end{table*}

\subsubsection{Modular configuration}

Modular architecture and training configurations introduce flexibility along multiple dimensions as discussed in section \ref{sec:modularity}. Adaptability is one such crucial feature that is enabled via modularization. In particular, it facilitates efficient adaptation in three  ways. The first is modality/task-specific customization. When a new modality/task emerges, specialized modules can be developed/replaced and integrated with existing framework without involving extensive changes in the architecture. This significantly lowers the barriers to extending model capabilities. The second is granular finetuning. Rather than updating all parameters, smaller modules can be fine-tuned on new datasets based on relevance. This is more efficient and less prone to overfitting. The third is external knowledge injection. Modularity provides clear interfaces for integrating external knowledge sources via RAG to augment the model capability.

In essence, adaptability is a fundamental characteristic of a GMM that enables several positive impacts in terms of better transfer learning across different tasks and modalities. 
This implies that models can be used as a starting point for other related tasks/modality without requiring extensive retraining or fine-tuning. Various adaptation strategies explored in the existing GMMs are listed in Table \ref{Tab:adap-strat}.

\begin{table*}
\caption{Generalist multimodal models description }
\label{Tab:pre-train-desc}
\scalebox{0.75}{
\begin{tabular}{|c|c|c|ccc|c|c|}
\hline
\multirow{2}{*}{\textbf{Model}} & {\textbf{Modalities}} & \multirow{2}{*}{\textbf{\begin{tabular}[c]{@{}c@{}}Training \\ objectives\end{tabular}}} & \multicolumn{3}{c|}{\textbf{Dataset}}                                                       & \multirow{2}{*}{\textbf{Code}} & \multirow{2}{*}{\textbf{\begin{tabular}[c]{@{}c@{}}Largest \\ model \end{tabular}}} \\ \cline{4-6}
                                &                                                                                              &  & \multicolumn{1}{c|}{\textbf{Text}}                                           & \multicolumn{1}{c|}{\textbf{Vision}}                                                                                     & \textbf{Cross}                                                                           & \textbf{}         \\ \hline
Uni-Perceiver                   & {\begin{tabular}[c]{@{}c@{}}Text \\ Images \\ Video \end{tabular}} & \begin{tabular}[c]{@{}c@{}}MLM, \\ Autoregressive, \\ NSP, \\
Image-text \\ retrieval, \\ MRM \end{tabular}                                                                                            & \multicolumn{1}{c|}{C4}                                                        & \multicolumn{1}{c|}{\begin{tabular}[c]{@{}c@{}}Imagenet, \\ OpenImages\\ Object365, \\ Imagenet-21k,\\ YFC100M\end{tabular}}                                                                                                    &  \multicolumn{1}{c|}{\begin{tabular}[c]{@{}c@{}}MiT, \\ Kinetics \end{tabular}}                                                                                        &  \href{https://github.com/fundamentalvision/Uni-Perceiver}{Github}   & 446M               \\ \hline
Unified-IO                      &  {\begin{tabular}[c]{@{}c@{}}Text \\ Images \\ Video \end{tabular}} & \begin{tabular}[c]{@{}c@{}}Text span- \\ denoising, \\ Mask image- \\ denoising \end{tabular}                                                                                            & \multicolumn{1}{c|}{\begin{tabular}[c]{@{}c@{}}C4,\\ Wiki \end{tabular} }                                                        & \multicolumn{1}{c|}{\begin{tabular}[c]{@{}c@{}}Imagenet-21k, \\ YFCC15M \\ CC12M \end{tabular}}                                                                                                   &   \multicolumn{1}{c|}{\begin{tabular}[c]{@{}c@{}}COCO-\\ caption, \\ VG-VQA, \\ SNLI-VE \end{tabular}}                                                                                           &       \href{https://unified-io.allenai.org/}{Website}        & 2.925B         \\ \hline
GATO                            &  {\begin{tabular}[c]{@{}c@{}}Text \\ Images \\ Pro-prioception \end{tabular}}  &                                          \begin{tabular}[c]{@{}c@{}} NSP \\ Supervised \\ label prediction, \\ reward maximization for \\ accomplishing goals \end{tabular}                                                 & \multicolumn{1}{c|}{\begin{tabular}[c]{@{}c@{}}Massive-\\ Text,\end{tabular}}                                                      & \multicolumn{1}{c|}{\begin{tabular}[c]{@{}c@{}}ImageNet,\\ \end{tabular}}                                                                                                  &                                                                        \multicolumn{1}{c|}{\begin{tabular}[c]{@{}c@{}}MSCOCO \\ Habitat-simulator\\ caption, \\ VQAV2, \\ M3W \end{tabular}}                     &                N/A  & 1.18B    \\ \hline
OFA+                             & {\begin{tabular}[c]{@{}c@{}}Text \\ Images \\ Video \\ Audio \\ Motion \\ Structural language \end{tabular}} & \begin{tabular}[c]{@{}c@{}} Span MLM, \\ Multimodal, \\ Instruction, \end{tabular}        & \multicolumn{1}{c|}{Pile}                                                    & \multicolumn{1}{c|}{\begin{tabular}[c]{@{}c@{}}COCO, \\ OpenImages\\ Object365, \\ Imagenet-21k,\\ YFC100M\end{tabular}} & \begin{tabular}[c]{@{}c@{}}CC3M, \\ VG-Cap,\\ VG-QA, \\ RefCOCO,\\ SBU, GQA\end{tabular} &  \href{https://github.com/OFA-Sys/OFA}{Github}           & 455M      \\ \hline
mPLUG-2                         & {\begin{tabular}[c]{@{}c@{}}Text \\ Images \\ Video \\ Audio \end{tabular}} & \begin{tabular}[c]{@{}c@{}}MLM, \\ Multimodal, \\ Instruction, \end{tabular}        & \multicolumn{1}{c|}{\begin{tabular}[c]{@{}c@{}}C4,\\ Wiki- \\ Corpus,\end{tabular}} & \multicolumn{1}{c|}{\begin{tabular}[c]{@{}c@{}} COCO,\\ Visual Genome \end{tabular}}                                          &{\begin{tabular}[c]{@{}c@{}} WebVid-2M, \\ Conceptual Captions\end{tabular}}                                                                          & \href{https://github.com/X-PLUG/mPLUG-2}{Github}    & 0.9B             \\ \hline
Meta-Transformer                         & {\begin{tabular}[c]{@{}c@{}}Text \\ Images \\ Video \\ Audio \\ Point cloud \\ Infrared \\ Hyper-spectrum \\ IMU \\ Graph \\ Time series \end{tabular}} &  \begin{tabular}[c]{@{}c@{}}Contrastive, \\ Learning \end{tabular}        & \multicolumn{1}{c|}{\begin{tabular}[c]{@{}c@{}}- \end{tabular}} & \multicolumn{1}{c|}{\begin{tabular}[c]{@{}c@{}}- \end{tabular}}                                                                                               & {\begin{tabular}[c]{@{}c@{}} LAION-2B \end{tabular}}                                                                               & \href{https://github.com/invictus717/MetaTransformer}{Github}  & 302M             \\ \hline
NExT-GPT                        & {\begin{tabular}[c]{@{}c@{}}Text \\ Images \\ Video \\ Audio  \end{tabular}} &  \begin{tabular}[c]{@{}c@{}}modality-switching  \\ instruction \\ tuning \end{tabular}        & \multicolumn{1}{c|}{\begin{tabular}[c]{@{}c@{}}-\end{tabular}} & \multicolumn{1}{c|}{\begin{tabular}[c]{@{}c@{}}- \end{tabular}}                                                                   & {\begin{tabular}[c]{@{}c@{}} MosIT \\ MS-COCO \\ CC3M \\ LAION \\ WebVid \\ AcitivityNet \\ VGGSS \end{tabular}}                                                                               & \href{https://next-gpt.github.io/}{Github}  & 7B             \\ \hline
OneLLM                       & {\begin{tabular}[c]{@{}c@{}}Text \\ Images \\ Video \\ Audio \\ Point cloud \\ IMU \\ Depth  \end{tabular}} &  \begin{tabular}[c]{@{}c@{}}Generative alignment \\ (BLIP2) \end{tabular}        & \multicolumn{1}{c|}{\begin{tabular}[c]{@{}c@{}}-\end{tabular}} & \multicolumn{1}{c|}{\begin{tabular}[c]{@{}c@{}}- \end{tabular}}                                                       & {\begin{tabular}[c]{@{}c@{}} LAION-2B   \\ LLaVA-150K \\ COCO Caption \\ VQAv2, OKVQA \\ Visual
Genome \\ MSRVTT-QA \\ AudioCaps \\ Ego4D, NSD \end{tabular}}                                                                               & \href{https://github.com/csuhan/OneLLM}{Github}  & 7B             \\ \hline
\end{tabular}
}
\end{table*}

\section{Challenges}
\label{sect-challenges}
\subsection{Underexplored Modalities}
Data deficiency is a serious challenge for GMMs because these models inherently require a large amount of data to be effectively trained. Multimodal datasets are limited in comparison to single modality language and vision datasets due to two factors, data generation cost and privacy concerns. Along with the standard data generation process, multimodal data also involves a systematic alignment among the modalities, which is a compute intensive task. As an example, the Multimodal Lecture Presentations (MLP) benchmark dataset involves hundreds of MTurkers to align and annotate educational lecture videos, slides, and audio signals \cite{lee2022multimodal}. With regard to privacy,  generation of realistic and diverse multimodal data often requires an enormous volume of training data, which may contain sensitive information. For example, in biomedical applications, multimodal data usually contains sensitive information about patients, such as medical records, radiology images, and genetic data. This sensitive information must be protected to ensure patient privacy and compliance with regulations such as HIPAA \cite{zhou2023comprehensive}. There are some initiatives in developing multimodal datasets (beyond text and image) as can be seen in NEXT-GPT~\citep{wu2023next} and ONELLM~\citep{han2023onellm}.
However, significant work remains in developing a systematic framework for multimodal data generation.

Furthermore, lack of data in novel modalities such as 3D point cloud or infrared images have led to under exploration; thus, critical challenges in these modalities remain unaddressed. This is evident in Table \ref{Tab:pre-train-desc} where only few models have the capability to handle these novel modalities.
There is a need to accommodate under explored modalities and their challenges. One such instance can be seen in  META-TRANSFORMER~\citep{zhang2023meta}, which captures thermal information of infrared images by encoding temperature values alongside visual features. This customized encoding drastically improves their downstream performance. However, additional, similar innovations across non-traditional modalities are needed.

\subsection{Weak Multimodal Evaluation}
Multimodal models are evaluated using multimodal benchmarks, datasets designed to test the generalization of models in multitask and transfer settings. In language and vision, there are several established benchmarks, including image-text retrieval and image-text question answering. Beyond these two modalities, few exist apart from those contained in MultiBench \cite{liang2021multibench}. Furthermore, commonly used evaluation metrics, such as accuracy, precision, recall, F1 score, and AUC-ROC, may not be sufficient for evaluating the performance of GMMs since they do not capture the complexity of interactions between modalities. In a multimodal setting, the dependencies between modalities and their relationship to tasks make it challenging to use traditional evaluation metrics. Therefore, it is necessary to devise more sophisticated evaluation metrics that can account for the unique challenges posed by multimodal data sources. In addition to expanding and diversifying the benchmark, a holistic view in terms of fairness, uncertainty estimation, and robustness, among other qualities, is also needed. A number of challenges with regard to existing benchmarks are highlighted in \cite{liang2022holistic}, and the authors also propose a novel benchmarking approach for GMMs. Inspiration should be drawn from \cite{liang2022holistic} while developing multimodal benchmarks.  

\subsection{Lack of Theoretical Multimodal Understanding}
GMMs are highly complex in regard to two factors, multiple heterogeneous modalities and large scale with millions/billions of parameters. This complexity makes it difficult to understand how these models work, how modalities interact, and what factors contribute to their performance. There has been some effort towards quantifying the heterogeneity of modalities, their interaction, and the consequent impact on model performance \cite{lianghighmmt,liang2022highmmt}. HIGHMMT \cite{lianghighmmt} proposes an information-theoretic approach called modality information transfer to quantify heterogeneity in modality. This approach uses a mutual information metric to measure the amount of transferable usable information from one modality to another, reflecting their similarity and interaction. Thereafter, the metric is employed to dynamically group modalities with similar characteristics for parameter sharing in the model.
However, there is still a long way to go in developing theoretical frameworks for a holistic understanding of multimodal mechanisms.
Moreover, many existing GMMs rely on heuristic techniques and ad-hoc design choices rather than being grounded on established theoretical concepts. This unsystematic practice can make it challenging to interpret the behavior of models or generalize results to new modalities or applications. Together, these shortcomings result in ineffective architectures and pretraining objectives.
In addition, there is a lack of cross-modality understanding in terms of transforming modality specific embeddings into language space. It remains a challenge to leverage prior knowledge from one modality/task to benefit the others. Thus, existing GMMs have limited extensibility to new modalities and involve expensive training costs. 

% \subsection{Poor Adaptability}
% Knowledge gain from one modality/task is not easily transferable to new modality/tasks

% \subsection{Weak performance on Science benchmarks}
% Most of the GMM are not evaluated rigorously for scientific tasks such as scientific captioning, chart reading, video understanding. 

\subsection{Absence of Multimodal Trust Framework} Most  GMMs prioritize predictive performance with little to no attention to trust assurance. Quantifying uncertainty is essential for safety-critical applications such as biomedical and hazard analysis \cite{verma2022attention, diemert2023can}. Accompanying predictions with confidence scores would enable end users to invoke informed decisions rather than blindly relying on model predictions. Recently, the FLM community began building uncertainty quantification tools based on standard paradigms including ensembles, Bayesian, and Monte-Carlo dropout techniques \cite{jiang2020can, xiao2022uncertainty,  kuhn2023semantic, lin2023generating, wagle2023empirical}. Similarly, in language-vision, UQ works such as KNOWNO~\citep{ren2023robots} leverage a conformal prediction framework to capture model uncertainty in robotic planning applications. However, there are little to no UQ efforts in GMMs space beyond language and image. Due to the large degree of data heterogeneity and multiple sub-components in GMMs architecture, it is critical to study the uncertainty across different modalities, tasks, and components so that customized evaluation approaches can be developed.

\subsection{Misaligned Modality Encodings}
Modality-specific encoders (Frozen) in GMMs are typically trained separately, leading to potential discrepancies in their representations (embeddings). This causes a heavy reliance on the projector/alignment module for effective learning. Therefore, designing a modality-shared encoder to encode multiple modalities remains a significant challenge. To reduce the misalignment in modality representations, there is a growing interest in developing homogenized encoders that would be trained to handle several modalities, such as  IMAGEBIND~\citep{girdhar2023imagebind}. Currently, it is limited to only six modalities. 

\subsection{High Computational Effort}
GMMs are generally compute intensive mainly due to their large size. However, apart from standard computational efforts, multimodality brings several new bottlenecks:
(i) GMMs typically involve processing high-dimensional input data, such as images and videos, along with the text. This requires huge memory with specialized hardware or distributed computing resources.
(ii) Integrating knowledge from several modalities requires careful data alignment. This can be challenging when the modalities have different scales, resolutions, and/or formats. For instance, alignment of slides, audio, and video in the Multimodal Lecture Presentations dataset involves several hours of human effort, as shown in \cite{lee2022multimodal}.
(iii) No explicit mechanism exists to assess cross modal/task performance and design robust pretext tasks without large scale training/evaluation.
There is a need to develop training strategies and architecture simplifications to reduce the computational effort in GMM development. For instance, rather than full-scale alignment between each diffusion decoder model and the universal LLM, NEXT-GPT~\citep{wu2023next} explores a decoding-side instruction-following alignment, which significantly reduces computation overhead. Similarly, NEXT-GPT's~\citep{wu2023next} architectural design samples visual tokens into fixed-length tokens and leverages unified encoders to significantly contribute in reducing computational effort of GMMs.

\section{Future opportunities}
\label{sect-opportunities}

\subsection{Expanding modality }
The fundamental bottleneck with the diverse domain of multimodal research is the lack of sufficient multimodal data in modalities outside of language and images. Therefore, there needs to be more emphasis on collecting/curating new datasets and benchmarks, especially in less explored modalities such as time series, audio, graphs, video, etc. Typically, alignments between modalities were done manually (e.g., crowdsourced). However, vision-language models have the capability to create multimodal datasets in terms of textual descriptions or captions for existing images or videos, adding a textual modality to the vision dataset. This generated data aligns with the content of the visual data; thuss contributing to the creation of multimodal datasets. Along with raw datasets for training, there is a serious gap in instruction-tuning multimodal data. A few exceptions include the MACAW-LLM \cite{lyu2023macaw}, DataComp \cite{gadre2023datacomp}, and the modality-switching instruction tuning dataset~\cite{wu2023next}.

% \subsection{Effective modular components}
% GMM constitutes various modular components which are used in an off-the-shelf manner. 

\subsection{Multimodal Prompting}

Most multimodal models are designed to generate text in response to text prompts, e.g. responding to user input through dialogue focused on a particular input modality~\cite{lyu2023macaw}.
Rather than using human-generated prompts, automatically generated prompts from pretrained models such as GPT can provide a wider variety of inputs for multimodal model learning, such as reinforcement learning to align the generated prompts with the target modality~\cite{yu2023fusing}.
Building on the success of prompt-tuning with text-only models, prompt-tuning with multimodal models can achieve high performance and more robustness to input variance with minimal extra training~\cite{yang2022prompt}. For instance, the diffusion model M-VADER~\citep{weinbach2022m} can generate new image or variations of the image based on the context input images and task description.
Accurately processing multimodal prompts may be difficult without a language model that is adapted to the target modality. In other words, simply using the most advanced text-only encoder model to handle prompts may be inadequate for modalities that are poorly understood or hard to describe with text, like seismic data.
Similarly, prompts that are outside the scope of a given modality may prove impossible for a model to handle and such impossible prompts may require a default ``I don't know'' response. This could occur in cases where the prompt is misaligned~\cite{rajpurkar2018know}, such as asking about audio in a static image.

% \subsection{Effective Unification}
% Explore different modality/task unification strategies beyond single transformer-based module.

\subsection{Extensive Scaling with Better Modular Components}
Most GMMs have not been studied under large scale, including both the size of a universal LLM and projector module (size of the largest configuration for each GMM is tabulated in Table.\ref{Tab:pre-train-desc}). This prevents us from exploiting the emerging abilities that are prominent in high scale. Recent efforts in FLMs have demonstrated that scaling offers a number of advantages, including higher model performance, better contextual comprehension, and better domain adaption through fine-tuning~\cite{kaplan2020scaling,hoffmann2022training}. Specifically, \cite{lu2023empirical} shows that scaling the size of the universal model (i.e.,LLM) in a multimodal setting significantly increases the performance in downstream tasks. There is scope to conduct such studies not only with respect to language models but also in regard to other multimodal components, including modality encoder, multimodal projector, and output decoder.

In the current setting, very standard components are leveraged in GMM training which may not be effective for downstream tasks. For instance, most  GMMs involve CLIP~\cite{radford2021learning} as the visual encoder. However, there is a need for research to improve modality encoders (visual/audio etc.) that better align with the given universal language model.
Furthermore,
inspired by the success of the unified multimodal encoder Imagebind~\cite{girdhar2023imagebind}, it would be interesting to explore the architecture of a unified decoder that can decode representations into any desirable modality. Another potential direction of enhancing GMMs is the multimodal retrieval augmented generation framework. Here, for the given multimodal query, relevant multimodal inputs can be first retrieved from the corpus, which will be fed to the universal multimodal model to further support it in the underlying tasks.

\subsection{Human Model Interaction}
With access to multiple modalities, GMMs can act as master module to integrate different learning components to understand and interpret human behavior more comprehensively. For instance, GMMs can analyze verbal and non-verbal cues (such as facial expressions, gestures, and tone of voice) to provide deeper insights into sentiment analysis, emotion recognition, and behavioral prediction. However, very little has been explored in this direction. In addition, human users may expect multimodal models to engage with them in a dialogue about multiple data inputs (e.g. ``compare these two images''), and this task may prove difficult if the model lacks extra provisions to store memory-intensive data such as video. In the space of model trust, it remains unclear how readily human users will tolerate errors from multimodal models, particularly hallucination errors, where the model appears incapable of seemingly simple tasks, such as identifying objects in an image. Following ongoing trends around AI explainability~\cite{stepin2021survey}, human users will likely seek increasingly detailed explanations for decisions made by multimodal models, such as a combination of natural language and attention maps over input~\cite{li2023joint}.

% \subsection{Enhancing Zero shot/few shot capability}
%  Instruction based pre/fine tuning enhance zero/few shot capability. Modality/task aware automatic Prompt designing is another promising technique for improving the performance of GMM. 

% Another direction of enabling zero-shot capabilities of GMM is to integrate retrieval based predictions with vector databases. Retrieval based LLM (R-LLM) first builds an index of the data/documents that it can retrieve. Then, given the query, R-LLM performs a retrieval step to identify the most relevant documents or responses from the index. The retrieved content along with query is then passed to the standard LM to generate response. The key advantage of R-LLM is their ability to work with any new index (database), thus offering a huge generalizability. 

\subsection{Emerging Capabilities}
There are various techniques that have not been explored much with respect to the training of GMMs. First, there is the impact of modality sequence during pretraining in the downstream tasks. There are plenty of research questions in this direction such as whether the sequence "text to image to video" or "image-video-text" performs better. In this regard, curriculum learning can also be explored. Second, how can we quantify the heterogenity level of the modality and how will the heterogeneity level impact the modality alignment in the representation space and, consequently, the downstream performance? Authors in \cite{liang2022highmmt} have done some preliminary analysis in this direction but are mostly focused on conventional architectures. In addition to various training techniques, one of the other emerging capabilities of FLMs is the ability to use them as knowledge modules to fuse external knowledge in smaller models designed for specific task/domain applications. This idea in some way aligns with the concept of world models where modules with specific knowledge are used to simulate scenarios \cite{matsuo2022deep}. For example, in the Voyager \cite{wang2023voyager}, an LLM is used as a world model to generate consistent action plans/policies for life-long learning in embodied tasks involving robotics and games. In this regard, GMMs would be a strong candidate for world models and can span wide applications ranging from image understanding tasks to reinforcement learning techniques for precise control actions. These are just few examples of emerging capabilities but GMMs will require lots of new work to understand and build next generation models that can perform tasks beyond our imagination.

\section{Summary and conclusions}
\label{sect-conclusion}
This paper reviews the generalist multimodal models that can work across multiple modalities beyond text and image. There are a series of new architecture and training schemes that have been consistently adopted in GMMs. Based on these widely adopted architectural features and training practices, we propose a novel taxonomy (unifiability, modularity, adaptability) to systematically analyze and categorize current GMMs. Through this taxonomic lens, we highlight unique design choices and their implication in model performances. We identify key challenges in GMMs space, including weak multimodal benchmarks and low scaling.
% We outline promising research directions to further simulate progress such as multimodal prompting and emerging capabilities.
This comprehensive review highlights unifying patterns and pressing open problems in the fast growing domain of GMMs. By condensing insights from relevant GMM works, this review will help multimodal community to develop more sophisticated GMMs through elegant architectural and training configurations.

\section*{Acknowledgements}
This work was supported by the NNSA Office of Defense Nuclear Nonproliferation Research and Development, U.S. Department of Energy, and Pacific Northwest National Laboratory, which is operated by Battelle Memorial Institute for the U.S. Department of Energy under Contract DE-AC05–76RLO1830. This article has been cleared
by PNNL for public release as PNNL-SA-198752.

\bibliographystyle{ACM-Reference-Format}
\bibliography{references}


\begin{thebibliography}{106}

%%% the Web address.

\ifx \showCODEN    \undefined \def \showCODEN     #1{\unskip}     \fi
\ifx \showDOI      \undefined \def \showDOI       #1{#1}\fi
\ifx \showISBNx    \undefined \def \showISBNx     #1{\unskip}     \fi
\ifx \showISBNxiii \undefined \def \showISBNxiii  #1{\unskip}     \fi
\ifx \showISSN     \undefined \def \showISSN      #1{\unskip}     \fi
\ifx \showLCCN     \undefined \def \showLCCN      #1{\unskip}     \fi
\ifx \shownote     \undefined \def \shownote      #1{#1}          \fi
\ifx \showarticletitle \undefined \def \showarticletitle #1{#1}   \fi
\ifx \showURL      \undefined \def \showURL       {\relax}        \fi
% invisible to TeX
\providecommand\bibfield[2]{#2}
\providecommand\bibinfo[2]{#2}
\providecommand\natexlab[1]{#1}
\providecommand\showeprint[2][]{arXiv:#2}

\bibitem[Acosta et~al\mbox{.}(2022)]%
        {acosta2022multimodal}
\bibfield{author}{\bibinfo{person}{Juli{\'a}n~N Acosta},
  \bibinfo{person}{Guido~J Falcone}, \bibinfo{person}{Pranav Rajpurkar}, {and}
  \bibinfo{person}{Eric~J Topol}.} \bibinfo{year}{2022}\natexlab{}.
\newblock \showarticletitle{Multimodal biomedical AI}.
\newblock \bibinfo{journal}{\emph{Nature Medicine}} \bibinfo{volume}{28},
  \bibinfo{number}{9} (\bibinfo{year}{2022}), \bibinfo{pages}{1773--1784}.
\newblock


\bibitem[Aghajanyan et~al\mbox{.}(2022)]%
        {aghajanyan2022cm3}
\bibfield{author}{\bibinfo{person}{Armen Aghajanyan}, \bibinfo{person}{Bernie
  Huang}, \bibinfo{person}{Candace Ross}, \bibinfo{person}{Vladimir Karpukhin},
  \bibinfo{person}{Hu Xu}, \bibinfo{person}{Naman Goyal},
  \bibinfo{person}{Dmytro Okhonko}, \bibinfo{person}{Mandar Joshi},
  \bibinfo{person}{Gargi Ghosh}, \bibinfo{person}{Mike Lewis}, {et~al\mbox{.}}}
  \bibinfo{year}{2022}\natexlab{}.
\newblock \showarticletitle{Cm3: A causal masked multimodal model of the
  internet}.
\newblock \bibinfo{journal}{\emph{arXiv preprint arXiv:2201.07520}}
  (\bibinfo{year}{2022}).
\newblock


\bibitem[Akbari et~al\mbox{.}(2021)]%
        {akbari2021vatt}
\bibfield{author}{\bibinfo{person}{Hassan Akbari}, \bibinfo{person}{Liangzhe
  Yuan}, \bibinfo{person}{Rui Qian}, \bibinfo{person}{Wei-Hong Chuang},
  \bibinfo{person}{Shih-Fu Chang}, \bibinfo{person}{Yin Cui}, {and}
  \bibinfo{person}{Boqing Gong}.} \bibinfo{year}{2021}\natexlab{}.
\newblock \showarticletitle{Vatt: Transformers for multimodal self-supervised
  learning from raw video, audio and text}.
\newblock \bibinfo{journal}{\emph{Advances in Neural Information Processing
  Systems}}  \bibinfo{volume}{34} (\bibinfo{year}{2021}),
  \bibinfo{pages}{24206--24221}.
\newblock


\bibitem[Alayrac et~al\mbox{.}(2022)]%
        {alayrac2022flamingo}
\bibfield{author}{\bibinfo{person}{Jean-Baptiste Alayrac},
  \bibinfo{person}{Jeff Donahue}, \bibinfo{person}{Pauline Luc},
  \bibinfo{person}{Antoine Miech}, \bibinfo{person}{Iain Barr},
  \bibinfo{person}{Yana Hasson}, \bibinfo{person}{Karel Lenc},
  \bibinfo{person}{Arthur Mensch}, \bibinfo{person}{Katie Millican},
  \bibinfo{person}{Malcolm Reynolds}, {et~al\mbox{.}}}
  \bibinfo{year}{2022}\natexlab{}.
\newblock \showarticletitle{Flamingo: a visual language model for few-shot
  learning}.
\newblock \bibinfo{journal}{\emph{arXiv preprint arXiv:2204.14198}}
  (\bibinfo{year}{2022}).
\newblock


\bibitem[Bai et~al\mbox{.}(2022)]%
        {bai2022ofasys}
\bibfield{author}{\bibinfo{person}{Jinze Bai}, \bibinfo{person}{Rui Men},
  \bibinfo{person}{Hao Yang}, \bibinfo{person}{Xuancheng Ren},
  \bibinfo{person}{Kai Dang}, \bibinfo{person}{Yichang Zhang},
  \bibinfo{person}{Xiaohuan Zhou}, \bibinfo{person}{Peng Wang},
  \bibinfo{person}{Sinan Tan}, \bibinfo{person}{An Yang}, {et~al\mbox{.}}}
  \bibinfo{year}{2022}\natexlab{}.
\newblock \showarticletitle{OFASys: A Multi-Modal Multi-Task Learning System
  for Building Generalist Models}.
\newblock \bibinfo{journal}{\emph{arXiv preprint arXiv:2212.04408}}
  (\bibinfo{year}{2022}).
\newblock


\bibitem[Brown et~al\mbox{.}(2020)]%
        {brown2020language}
\bibfield{author}{\bibinfo{person}{Tom Brown}, \bibinfo{person}{Benjamin Mann},
  \bibinfo{person}{Nick Ryder}, \bibinfo{person}{Melanie Subbiah},
  \bibinfo{person}{Jared~D Kaplan}, \bibinfo{person}{Prafulla Dhariwal},
  \bibinfo{person}{Arvind Neelakantan}, \bibinfo{person}{Pranav Shyam},
  \bibinfo{person}{Girish Sastry}, \bibinfo{person}{Amanda Askell},
  {et~al\mbox{.}}} \bibinfo{year}{2020}\natexlab{}.
\newblock \showarticletitle{Language models are few-shot learners}.
\newblock \bibinfo{journal}{\emph{Advances in neural information processing
  systems}}  \bibinfo{volume}{33} (\bibinfo{year}{2020}),
  \bibinfo{pages}{1877--1901}.
\newblock


\bibitem[Chen et~al\mbox{.}(2021)]%
        {chen2021pix2seq}
\bibfield{author}{\bibinfo{person}{Ting Chen}, \bibinfo{person}{Saurabh
  Saxena}, \bibinfo{person}{Lala Li}, \bibinfo{person}{David~J Fleet}, {and}
  \bibinfo{person}{Geoffrey Hinton}.} \bibinfo{year}{2021}\natexlab{}.
\newblock \showarticletitle{Pix2seq: A language modeling framework for object
  detection}.
\newblock \bibinfo{journal}{\emph{arXiv preprint arXiv:2109.10852}}
  (\bibinfo{year}{2021}).
\newblock


\bibitem[Chiang et~al\mbox{.}(2023)]%
        {chiang2023vicuna}
\bibfield{author}{\bibinfo{person}{Wei-Lin Chiang}, \bibinfo{person}{Zhuohan
  Li}, \bibinfo{person}{Zi Lin}, \bibinfo{person}{Ying Sheng},
  \bibinfo{person}{Zhanghao Wu}, \bibinfo{person}{Hao Zhang},
  \bibinfo{person}{Lianmin Zheng}, \bibinfo{person}{Siyuan Zhuang},
  \bibinfo{person}{Yonghao Zhuang}, \bibinfo{person}{Joseph~E Gonzalez},
  {et~al\mbox{.}}} \bibinfo{year}{2023}\natexlab{}.
\newblock \showarticletitle{Vicuna: An open-source chatbot impressing gpt-4
  with 90\%* chatgpt quality}.
\newblock \bibinfo{journal}{\emph{See https://vicuna. lmsys. org (accessed 14
  April 2023)}} (\bibinfo{year}{2023}).
\newblock


\bibitem[Chung et~al\mbox{.}(2022)]%
        {chung2022scaling}
\bibfield{author}{\bibinfo{person}{Hyung~Won Chung}, \bibinfo{person}{Le Hou},
  \bibinfo{person}{Shayne Longpre}, \bibinfo{person}{Barret Zoph},
  \bibinfo{person}{Yi Tay}, \bibinfo{person}{William Fedus},
  \bibinfo{person}{Eric Li}, \bibinfo{person}{Xuezhi Wang},
  \bibinfo{person}{Mostafa Dehghani}, \bibinfo{person}{Siddhartha Brahma},
  {et~al\mbox{.}}} \bibinfo{year}{2022}\natexlab{}.
\newblock \showarticletitle{Scaling instruction-finetuned language models}.
\newblock \bibinfo{journal}{\emph{arXiv preprint arXiv:2210.11416}}
  (\bibinfo{year}{2022}).
\newblock


\bibitem[Dao et~al\mbox{.}(2022)]%
        {dao2022flashattention}
\bibfield{author}{\bibinfo{person}{Tri Dao}, \bibinfo{person}{Dan Fu},
  \bibinfo{person}{Stefano Ermon}, \bibinfo{person}{Atri Rudra}, {and}
  \bibinfo{person}{Christopher R{\'e}}.} \bibinfo{year}{2022}\natexlab{}.
\newblock \showarticletitle{Flashattention: Fast and memory-efficient exact
  attention with io-awareness}.
\newblock \bibinfo{journal}{\emph{Advances in Neural Information Processing
  Systems}}  \bibinfo{volume}{35} (\bibinfo{year}{2022}),
  \bibinfo{pages}{16344--16359}.
\newblock


\bibitem[Das et~al\mbox{.}(2023)]%
        {das2023there}
\bibfield{author}{\bibinfo{person}{Laya Das}, \bibinfo{person}{Sai Munikoti},
  {and} \bibinfo{person}{Mahantesh Halappanavar}.}
  \bibinfo{year}{2023}\natexlab{}.
\newblock \showarticletitle{There is more to graphs than meets the eye:
  Learning universal features with self-supervision}.
\newblock \bibinfo{journal}{\emph{arXiv preprint arXiv:2305.19871}}
  (\bibinfo{year}{2023}).
\newblock


\bibitem[Devlin et~al\mbox{.}(2018)]%
        {devlin2018bert}
\bibfield{author}{\bibinfo{person}{Jacob Devlin}, \bibinfo{person}{Ming-Wei
  Chang}, \bibinfo{person}{Kenton Lee}, {and} \bibinfo{person}{Kristina
  Toutanova}.} \bibinfo{year}{2018}\natexlab{}.
\newblock \showarticletitle{Bert: Pre-training of deep bidirectional
  transformers for language understanding}.
\newblock \bibinfo{journal}{\emph{arXiv preprint arXiv:1810.04805}}
  (\bibinfo{year}{2018}).
\newblock


\bibitem[Diemert and Weber(2023)]%
        {diemert2023can}
\bibfield{author}{\bibinfo{person}{Simon Diemert} {and} \bibinfo{person}{Jens~H
  Weber}.} \bibinfo{year}{2023}\natexlab{}.
\newblock \showarticletitle{Can Large Language Models assist in Hazard
  Analysis?}
\newblock \bibinfo{journal}{\emph{arXiv preprint arXiv:2303.15473}}
  (\bibinfo{year}{2023}).
\newblock


\bibitem[Dosovitskiy et~al\mbox{.}(2020)]%
        {dosovitskiy2020image}
\bibfield{author}{\bibinfo{person}{Alexey Dosovitskiy}, \bibinfo{person}{Lucas
  Beyer}, \bibinfo{person}{Alexander Kolesnikov}, \bibinfo{person}{Dirk
  Weissenborn}, \bibinfo{person}{Xiaohua Zhai}, \bibinfo{person}{Thomas
  Unterthiner}, \bibinfo{person}{Mostafa Dehghani}, \bibinfo{person}{Matthias
  Minderer}, \bibinfo{person}{Georg Heigold}, \bibinfo{person}{Sylvain Gelly},
  {et~al\mbox{.}}} \bibinfo{year}{2020}\natexlab{}.
\newblock \showarticletitle{An image is worth 16x16 words: Transformers for
  image recognition at scale}.
\newblock \bibinfo{journal}{\emph{arXiv preprint arXiv:2010.11929}}
  (\bibinfo{year}{2020}).
\newblock


\bibitem[Esser et~al\mbox{.}(2021)]%
        {esser2021taming}
\bibfield{author}{\bibinfo{person}{Patrick Esser}, \bibinfo{person}{Robin
  Rombach}, {and} \bibinfo{person}{Bjorn Ommer}.}
  \bibinfo{year}{2021}\natexlab{}.
\newblock \showarticletitle{Taming transformers for high-resolution image
  synthesis}. In \bibinfo{booktitle}{\emph{Proceedings of the IEEE/CVF
  conference on computer vision and pattern recognition}}.
  \bibinfo{pages}{12873--12883}.
\newblock


\bibitem[Gadre et~al\mbox{.}(2023)]%
        {gadre2023datacomp}
\bibfield{author}{\bibinfo{person}{Samir~Yitzhak Gadre},
  \bibinfo{person}{Gabriel Ilharco}, \bibinfo{person}{Alex Fang},
  \bibinfo{person}{Jonathan Hayase}, \bibinfo{person}{Georgios Smyrnis},
  \bibinfo{person}{Thao Nguyen}, \bibinfo{person}{Ryan Marten},
  \bibinfo{person}{Mitchell Wortsman}, \bibinfo{person}{Dhruba Ghosh},
  \bibinfo{person}{Jieyu Zhang}, {et~al\mbox{.}}}
  \bibinfo{year}{2023}\natexlab{}.
\newblock \showarticletitle{DataComp: In search of the next generation of
  multimodal datasets}.
\newblock \bibinfo{journal}{\emph{arXiv preprint arXiv:2304.14108}}
  (\bibinfo{year}{2023}).
\newblock


\bibitem[Gao et~al\mbox{.}(2020)]%
        {gao2020survey}
\bibfield{author}{\bibinfo{person}{Jing Gao}, \bibinfo{person}{Peng Li},
  \bibinfo{person}{Zhikui Chen}, {and} \bibinfo{person}{Jianing Zhang}.}
  \bibinfo{year}{2020}\natexlab{}.
\newblock \showarticletitle{A survey on deep learning for multimodal data
  fusion}.
\newblock \bibinfo{journal}{\emph{Neural Computation}} \bibinfo{volume}{32},
  \bibinfo{number}{5} (\bibinfo{year}{2020}), \bibinfo{pages}{829--864}.
\newblock


\bibitem[Garza and Mergenthaler-Canseco(2023)]%
        {garza2023timegpt}
\bibfield{author}{\bibinfo{person}{Azul Garza} {and} \bibinfo{person}{Max
  Mergenthaler-Canseco}.} \bibinfo{year}{2023}\natexlab{}.
\newblock \showarticletitle{TimeGPT-1}.
\newblock \bibinfo{journal}{\emph{arXiv preprint arXiv:2310.03589}}
  (\bibinfo{year}{2023}).
\newblock


\bibitem[Girdhar et~al\mbox{.}(2023)]%
        {girdhar2023imagebind}
\bibfield{author}{\bibinfo{person}{Rohit Girdhar}, \bibinfo{person}{Alaaeldin
  El-Nouby}, \bibinfo{person}{Zhuang Liu}, \bibinfo{person}{Mannat Singh},
  \bibinfo{person}{Kalyan~Vasudev Alwala}, \bibinfo{person}{Armand Joulin},
  {and} \bibinfo{person}{Ishan Misra}.} \bibinfo{year}{2023}\natexlab{}.
\newblock \showarticletitle{Imagebind: One embedding space to bind them all}.
  In \bibinfo{booktitle}{\emph{Proceedings of the IEEE/CVF Conference on
  Computer Vision and Pattern Recognition}}. \bibinfo{pages}{15180--15190}.
\newblock


\bibitem[Goyal(2022)]%
        {goyal2022survey}
\bibfield{author}{\bibinfo{person}{Naman Goyal}.}
  \bibinfo{year}{2022}\natexlab{}.
\newblock \showarticletitle{A survey on Self Supervised learning approaches for
  improving Multimodal representation learning}.
\newblock \bibinfo{journal}{\emph{arXiv preprint arXiv:2210.11024}}
  (\bibinfo{year}{2022}).
\newblock


\bibitem[Grigsby et~al\mbox{.}(2021)]%
        {grigsby2021long}
\bibfield{author}{\bibinfo{person}{Jake Grigsby}, \bibinfo{person}{Zhe Wang},
  {and} \bibinfo{person}{Yanjun Qi}.} \bibinfo{year}{2021}\natexlab{}.
\newblock \showarticletitle{Long-range transformers for dynamic spatiotemporal
  forecasting}.
\newblock \bibinfo{journal}{\emph{arXiv preprint arXiv:2109.12218}}
  (\bibinfo{year}{2021}).
\newblock


\bibitem[Han et~al\mbox{.}(2023)]%
        {han2023onellm}
\bibfield{author}{\bibinfo{person}{Jiaming Han}, \bibinfo{person}{Kaixiong
  Gong}, \bibinfo{person}{Yiyuan Zhang}, \bibinfo{person}{Jiaqi Wang},
  \bibinfo{person}{Kaipeng Zhang}, \bibinfo{person}{Dahua Lin},
  \bibinfo{person}{Yu Qiao}, \bibinfo{person}{Peng Gao}, {and}
  \bibinfo{person}{Xiangyu Yue}.} \bibinfo{year}{2023}\natexlab{}.
\newblock \showarticletitle{OneLLM: One Framework to Align All Modalities with
  Language}.
\newblock \bibinfo{journal}{\emph{arXiv preprint arXiv:2312.03700}}
  (\bibinfo{year}{2023}).
\newblock


\bibitem[Hoffmann et~al\mbox{.}(2022)]%
        {hoffmann2022training}
\bibfield{author}{\bibinfo{person}{Jordan Hoffmann}, \bibinfo{person}{Sebastian
  Borgeaud}, \bibinfo{person}{Arthur Mensch}, \bibinfo{person}{Elena
  Buchatskaya}, \bibinfo{person}{Trevor Cai}, \bibinfo{person}{Eliza
  Rutherford}, \bibinfo{person}{Diego de~Las Casas}, \bibinfo{person}{Lisa~Anne
  Hendricks}, \bibinfo{person}{Johannes Welbl}, \bibinfo{person}{Aidan Clark},
  {et~al\mbox{.}}} \bibinfo{year}{2022}\natexlab{}.
\newblock \showarticletitle{Training compute-optimal large language models}.
\newblock \bibinfo{journal}{\emph{arXiv preprint arXiv:2203.15556}}
  (\bibinfo{year}{2022}).
\newblock


\bibitem[Hofst{\"a}tter et~al\mbox{.}(2023)]%
        {hofstatter2023fid}
\bibfield{author}{\bibinfo{person}{Sebastian Hofst{\"a}tter},
  \bibinfo{person}{Jiecao Chen}, \bibinfo{person}{Karthik Raman}, {and}
  \bibinfo{person}{Hamed Zamani}.} \bibinfo{year}{2023}\natexlab{}.
\newblock \showarticletitle{Fid-light: Efficient and effective
  retrieval-augmented text generation}. In
  \bibinfo{booktitle}{\emph{Proceedings of the 46th International ACM SIGIR
  Conference on Research and Development in Information Retrieval}}.
  \bibinfo{pages}{1437--1447}.
\newblock


\bibitem[Horawalavithana et~al\mbox{.}(2023)]%
        {horawalavithana2023scitune}
\bibfield{author}{\bibinfo{person}{Sameera Horawalavithana},
  \bibinfo{person}{Sai Munikoti}, \bibinfo{person}{Ian Stewart}, {and}
  \bibinfo{person}{Henry Kvinge}.} \bibinfo{year}{2023}\natexlab{}.
\newblock \showarticletitle{Scitune: Aligning large language models with
  scientific multimodal instructions}.
\newblock \bibinfo{journal}{\emph{arXiv preprint arXiv:2307.01139}}
  (\bibinfo{year}{2023}).
\newblock


\bibitem[Hu et~al\mbox{.}(2021)]%
        {hu2021lora}
\bibfield{author}{\bibinfo{person}{Edward~J Hu}, \bibinfo{person}{Yelong Shen},
  \bibinfo{person}{Phillip Wallis}, \bibinfo{person}{Zeyuan Allen-Zhu},
  \bibinfo{person}{Yuanzhi Li}, \bibinfo{person}{Shean Wang},
  \bibinfo{person}{Lu Wang}, {and} \bibinfo{person}{Weizhu Chen}.}
  \bibinfo{year}{2021}\natexlab{}.
\newblock \showarticletitle{Lora: Low-rank adaptation of large language
  models}.
\newblock \bibinfo{journal}{\emph{arXiv preprint arXiv:2106.09685}}
  (\bibinfo{year}{2021}).
\newblock


\bibitem[Hu et~al\mbox{.}(2023)]%
        {hu2023reveal}
\bibfield{author}{\bibinfo{person}{Ziniu Hu}, \bibinfo{person}{Ahmet Iscen},
  \bibinfo{person}{Chen Sun}, \bibinfo{person}{Zirui Wang},
  \bibinfo{person}{Kai-Wei Chang}, \bibinfo{person}{Yizhou Sun},
  \bibinfo{person}{Cordelia Schmid}, \bibinfo{person}{David~A Ross}, {and}
  \bibinfo{person}{Alireza Fathi}.} \bibinfo{year}{2023}\natexlab{}.
\newblock \showarticletitle{Reveal: Retrieval-augmented visual-language
  pre-training with multi-source multimodal knowledge memory}. In
  \bibinfo{booktitle}{\emph{Proceedings of the IEEE/CVF Conference on Computer
  Vision and Pattern Recognition}}. \bibinfo{pages}{23369--23379}.
\newblock


\bibitem[Huang et~al\mbox{.}(2023)]%
        {huang2023language}
\bibfield{author}{\bibinfo{person}{Shaohan Huang}, \bibinfo{person}{Li Dong},
  \bibinfo{person}{Wenhui Wang}, \bibinfo{person}{Yaru Hao},
  \bibinfo{person}{Saksham Singhal}, \bibinfo{person}{Shuming Ma},
  \bibinfo{person}{Tengchao Lv}, \bibinfo{person}{Lei Cui},
  \bibinfo{person}{Owais~Khan Mohammed}, \bibinfo{person}{Qiang Liu},
  {et~al\mbox{.}}} \bibinfo{year}{2023}\natexlab{}.
\newblock \showarticletitle{Language Is Not All You Need: Aligning Perception
  with Language Models}.
\newblock \bibinfo{journal}{\emph{arXiv preprint arXiv:2302.14045}}
  (\bibinfo{year}{2023}).
\newblock


\bibitem[Huang et~al\mbox{.}(2022)]%
        {huang2022modality}
\bibfield{author}{\bibinfo{person}{Yu Huang}, \bibinfo{person}{Junyang Lin},
  \bibinfo{person}{Chang Zhou}, \bibinfo{person}{Hongxia Yang}, {and}
  \bibinfo{person}{Longbo Huang}.} \bibinfo{year}{2022}\natexlab{}.
\newblock \showarticletitle{Modality competition: What makes joint training of
  multi-modal network fail in deep learning?(provably)}. In
  \bibinfo{booktitle}{\emph{International Conference on Machine Learning}}.
  PMLR, \bibinfo{pages}{9226--9259}.
\newblock


\bibitem[HuggingFace(2023)]%
        {zeroscope}
\bibfield{author}{\bibinfo{person}{HuggingFace}.}
  \bibinfo{year}{2023}\natexlab{}.
\newblock \bibinfo{title}{Cerspense. Zeroscope: Diffusion-based text-to-video
  synthesis.}
\newblock
\newblock
\urldef\tempurl%
\url{https://huggingface. co/cerspense.}
\showURL{%
Retrieved September 7, 2023 from \tempurl}


\bibitem[Huo et~al\mbox{.}(2021)]%
        {huo2021wenlan}
\bibfield{author}{\bibinfo{person}{Yuqi Huo}, \bibinfo{person}{Manli Zhang},
  \bibinfo{person}{Guangzhen Liu}, \bibinfo{person}{Haoyu Lu},
  \bibinfo{person}{Yizhao Gao}, \bibinfo{person}{Guoxing Yang},
  \bibinfo{person}{Jingyuan Wen}, \bibinfo{person}{Heng Zhang},
  \bibinfo{person}{Baogui Xu}, \bibinfo{person}{Weihao Zheng}, {et~al\mbox{.}}}
  \bibinfo{year}{2021}\natexlab{}.
\newblock \showarticletitle{WenLan: Bridging vision and language by large-scale
  multi-modal pre-training}.
\newblock \bibinfo{journal}{\emph{arXiv preprint arXiv:2103.06561}}
  (\bibinfo{year}{2021}).
\newblock


\bibitem[Izacard et~al\mbox{.}(2022)]%
        {izacard2022few}
\bibfield{author}{\bibinfo{person}{Gautier Izacard}, \bibinfo{person}{Patrick
  Lewis}, \bibinfo{person}{Maria Lomeli}, \bibinfo{person}{Lucas Hosseini},
  \bibinfo{person}{Fabio Petroni}, \bibinfo{person}{Timo Schick},
  \bibinfo{person}{Jane Dwivedi-Yu}, \bibinfo{person}{Armand Joulin},
  \bibinfo{person}{Sebastian Riedel}, {and} \bibinfo{person}{Edouard Grave}.}
  \bibinfo{year}{2022}\natexlab{}.
\newblock \showarticletitle{Few-shot learning with retrieval augmented language
  models}.
\newblock \bibinfo{journal}{\emph{arXiv preprint arXiv:2208.03299}}
  (\bibinfo{year}{2022}).
\newblock


\bibitem[Jaegle et~al\mbox{.}(2021a)]%
        {jaegle2021perceiverio}
\bibfield{author}{\bibinfo{person}{Andrew Jaegle}, \bibinfo{person}{Sebastian
  Borgeaud}, \bibinfo{person}{Jean-Baptiste Alayrac}, \bibinfo{person}{Carl
  Doersch}, \bibinfo{person}{Catalin Ionescu}, \bibinfo{person}{David Ding},
  \bibinfo{person}{Skanda Koppula}, \bibinfo{person}{Daniel Zoran},
  \bibinfo{person}{Andrew Brock}, \bibinfo{person}{Evan Shelhamer},
  {et~al\mbox{.}}} \bibinfo{year}{2021}\natexlab{a}.
\newblock \showarticletitle{Perceiver io: A general architecture for structured
  inputs \& outputs}.
\newblock \bibinfo{journal}{\emph{arXiv preprint arXiv:2107.14795}}
  (\bibinfo{year}{2021}).
\newblock


\bibitem[Jaegle et~al\mbox{.}(2021b)]%
        {jaegle2021perceiver}
\bibfield{author}{\bibinfo{person}{Andrew Jaegle}, \bibinfo{person}{Felix
  Gimeno}, \bibinfo{person}{Andy Brock}, \bibinfo{person}{Oriol Vinyals},
  \bibinfo{person}{Andrew Zisserman}, {and} \bibinfo{person}{Joao Carreira}.}
  \bibinfo{year}{2021}\natexlab{b}.
\newblock \showarticletitle{Perceiver: General perception with iterative
  attention}. In \bibinfo{booktitle}{\emph{International conference on machine
  learning}}. PMLR, \bibinfo{pages}{4651--4664}.
\newblock


\bibitem[Jiang et~al\mbox{.}(2023)]%
        {jiang2023mistral}
\bibfield{author}{\bibinfo{person}{Albert~Q Jiang}, \bibinfo{person}{Alexandre
  Sablayrolles}, \bibinfo{person}{Arthur Mensch}, \bibinfo{person}{Chris
  Bamford}, \bibinfo{person}{Devendra~Singh Chaplot}, \bibinfo{person}{Diego
  de~las Casas}, \bibinfo{person}{Florian Bressand}, \bibinfo{person}{Gianna
  Lengyel}, \bibinfo{person}{Guillaume Lample}, \bibinfo{person}{Lucile
  Saulnier}, {et~al\mbox{.}}} \bibinfo{year}{2023}\natexlab{}.
\newblock \showarticletitle{Mistral 7B}.
\newblock \bibinfo{journal}{\emph{arXiv preprint arXiv:2310.06825}}
  (\bibinfo{year}{2023}).
\newblock


\bibitem[Jiang et~al\mbox{.}(2020)]%
        {jiang2020can}
\bibfield{author}{\bibinfo{person}{Zhengbao Jiang}, \bibinfo{person}{Frank~F
  Xu}, \bibinfo{person}{Jun Araki}, {and} \bibinfo{person}{Graham Neubig}.}
  \bibinfo{year}{2020}\natexlab{}.
\newblock \showarticletitle{How can we know what language models know?}
\newblock \bibinfo{journal}{\emph{Transactions of the Association for
  Computational Linguistics}}  \bibinfo{volume}{8} (\bibinfo{year}{2020}),
  \bibinfo{pages}{423--438}.
\newblock


\bibitem[Jin et~al\mbox{.}(2021a)]%
        {jin2021node}
\bibfield{author}{\bibinfo{person}{Wei Jin}, \bibinfo{person}{Tyler Derr},
  \bibinfo{person}{Yiqi Wang}, \bibinfo{person}{Yao Ma}, \bibinfo{person}{Zitao
  Liu}, {and} \bibinfo{person}{Jiliang Tang}.}
  \bibinfo{year}{2021}\natexlab{a}.
\newblock \showarticletitle{Node similarity preserving graph convolutional
  networks}. In \bibinfo{booktitle}{\emph{Proceedings of the 14th ACM
  international conference on web search and data mining}}.
  \bibinfo{pages}{148--156}.
\newblock


\bibitem[Jin et~al\mbox{.}(2021b)]%
        {jin2021automated}
\bibfield{author}{\bibinfo{person}{Wei Jin}, \bibinfo{person}{Xiaorui Liu},
  \bibinfo{person}{Xiangyu Zhao}, \bibinfo{person}{Yao Ma},
  \bibinfo{person}{Neil Shah}, {and} \bibinfo{person}{Jiliang Tang}.}
  \bibinfo{year}{2021}\natexlab{b}.
\newblock \showarticletitle{Automated self-supervised learning for graphs}.
\newblock \bibinfo{journal}{\emph{arXiv preprint arXiv:2106.05470}}
  (\bibinfo{year}{2021}).
\newblock


\bibitem[Kaplan et~al\mbox{.}(2020)]%
        {kaplan2020scaling}
\bibfield{author}{\bibinfo{person}{Jared Kaplan}, \bibinfo{person}{Sam
  McCandlish}, \bibinfo{person}{Tom Henighan}, \bibinfo{person}{Tom~B Brown},
  \bibinfo{person}{Benjamin Chess}, \bibinfo{person}{Rewon Child},
  \bibinfo{person}{Scott Gray}, \bibinfo{person}{Alec Radford},
  \bibinfo{person}{Jeffrey Wu}, {and} \bibinfo{person}{Dario Amodei}.}
  \bibinfo{year}{2020}\natexlab{}.
\newblock \showarticletitle{Scaling laws for neural language models}.
\newblock \bibinfo{journal}{\emph{arXiv preprint arXiv:2001.08361}}
  (\bibinfo{year}{2020}).
\newblock


\bibitem[Khattab et~al\mbox{.}(2022)]%
        {khattab2022demonstrate}
\bibfield{author}{\bibinfo{person}{Omar Khattab}, \bibinfo{person}{Keshav
  Santhanam}, \bibinfo{person}{Xiang~Lisa Li}, \bibinfo{person}{David Hall},
  \bibinfo{person}{Percy Liang}, \bibinfo{person}{Christopher Potts}, {and}
  \bibinfo{person}{Matei Zaharia}.} \bibinfo{year}{2022}\natexlab{}.
\newblock \showarticletitle{Demonstrate-Search-Predict: Composing retrieval and
  language models for knowledge-intensive NLP}.
\newblock \bibinfo{journal}{\emph{arXiv preprint arXiv:2212.14024}}
  (\bibinfo{year}{2022}).
\newblock


\bibitem[Kuhn et~al\mbox{.}(2023)]%
        {kuhn2023semantic}
\bibfield{author}{\bibinfo{person}{Lorenz Kuhn}, \bibinfo{person}{Yarin Gal},
  {and} \bibinfo{person}{Sebastian Farquhar}.} \bibinfo{year}{2023}\natexlab{}.
\newblock \showarticletitle{Semantic uncertainty: Linguistic invariances for
  uncertainty estimation in natural language generation}.
\newblock \bibinfo{journal}{\emph{arXiv preprint arXiv:2302.09664}}
  (\bibinfo{year}{2023}).
\newblock


\bibitem[Lee et~al\mbox{.}(2022)]%
        {lee2022multimodal}
\bibfield{author}{\bibinfo{person}{Dong~Won Lee}, \bibinfo{person}{Chaitanya
  Ahuja}, \bibinfo{person}{Paul~Pu Liang}, \bibinfo{person}{Sanika Natu}, {and}
  \bibinfo{person}{Louis-Philippe Morency}.} \bibinfo{year}{2022}\natexlab{}.
\newblock \showarticletitle{Multimodal Lecture Presentations Dataset:
  Understanding Multimodality in Educational Slides}.
\newblock \bibinfo{journal}{\emph{arXiv preprint arXiv:2208.08080}}
  (\bibinfo{year}{2022}).
\newblock


\bibitem[Li et~al\mbox{.}(2023a)]%
        {li2023multimodal}
\bibfield{author}{\bibinfo{person}{Chunyuan Li}, \bibinfo{person}{Zhe Gan},
  \bibinfo{person}{Zhengyuan Yang}, \bibinfo{person}{Jianwei Yang},
  \bibinfo{person}{Linjie Li}, \bibinfo{person}{Lijuan Wang}, {and}
  \bibinfo{person}{Jianfeng Gao}.} \bibinfo{year}{2023}\natexlab{a}.
\newblock \showarticletitle{Multimodal foundation models: From specialists to
  general-purpose assistants}.
\newblock \bibinfo{journal}{\emph{arXiv preprint arXiv:2309.10020}}
  (\bibinfo{year}{2023}).
\newblock


\bibitem[Li et~al\mbox{.}(2022)]%
        {li2022mplug}
\bibfield{author}{\bibinfo{person}{Chenliang Li}, \bibinfo{person}{Haiyang Xu},
  \bibinfo{person}{Junfeng Tian}, \bibinfo{person}{Wei Wang},
  \bibinfo{person}{Ming Yan}, \bibinfo{person}{Bin Bi}, \bibinfo{person}{Jiabo
  Ye}, \bibinfo{person}{Hehong Chen}, \bibinfo{person}{Guohai Xu},
  \bibinfo{person}{Zheng Cao}, {et~al\mbox{.}}}
  \bibinfo{year}{2022}\natexlab{}.
\newblock \showarticletitle{mPLUG: Effective and Efficient Vision-Language
  Learning by Cross-modal Skip-connections}.
\newblock \bibinfo{journal}{\emph{arXiv preprint arXiv:2205.12005}}
  (\bibinfo{year}{2022}).
\newblock


\bibitem[Li et~al\mbox{.}(2021)]%
        {li2021align}
\bibfield{author}{\bibinfo{person}{Junnan Li}, \bibinfo{person}{Ramprasaath
  Selvaraju}, \bibinfo{person}{Akhilesh Gotmare}, \bibinfo{person}{Shafiq
  Joty}, \bibinfo{person}{Caiming Xiong}, {and} \bibinfo{person}{Steven
  Chu~Hong Hoi}.} \bibinfo{year}{2021}\natexlab{}.
\newblock \showarticletitle{Align before fuse: Vision and language
  representation learning with momentum distillation}.
\newblock \bibinfo{journal}{\emph{Advances in neural information processing
  systems}}  \bibinfo{volume}{34} (\bibinfo{year}{2021}),
  \bibinfo{pages}{9694--9705}.
\newblock


\bibitem[Li et~al\mbox{.}(2023b)]%
        {li2023joint}
\bibfield{author}{\bibinfo{person}{Zhenyang Li}, \bibinfo{person}{Yangyang
  Guo}, \bibinfo{person}{Kejie Wang}, \bibinfo{person}{Yinwei Wei},
  \bibinfo{person}{Liqiang Nie}, {and} \bibinfo{person}{Mohan Kankanhalli}.}
  \bibinfo{year}{2023}\natexlab{b}.
\newblock \showarticletitle{Joint answering and explanation for visual
  commonsense reasoning}.
\newblock \bibinfo{journal}{\emph{IEEE Transactions on Image Processing}}
  (\bibinfo{year}{2023}).
\newblock


\bibitem[Liang et~al\mbox{.}(2022a)]%
        {liang2022holistic}
\bibfield{author}{\bibinfo{person}{Percy Liang}, \bibinfo{person}{Rishi
  Bommasani}, \bibinfo{person}{Tony Lee}, \bibinfo{person}{Dimitris Tsipras},
  \bibinfo{person}{Dilara Soylu}, \bibinfo{person}{Michihiro Yasunaga},
  \bibinfo{person}{Yian Zhang}, \bibinfo{person}{Deepak Narayanan},
  \bibinfo{person}{Yuhuai Wu}, \bibinfo{person}{Ananya Kumar}, {et~al\mbox{.}}}
  \bibinfo{year}{2022}\natexlab{a}.
\newblock \showarticletitle{Holistic evaluation of language models}.
\newblock \bibinfo{journal}{\emph{arXiv preprint arXiv:2211.09110}}
  (\bibinfo{year}{2022}).
\newblock


\bibitem[Liang et~al\mbox{.}(2022b)]%
        {liang2022highmmt}
\bibfield{author}{\bibinfo{person}{Paul~Pu Liang}, \bibinfo{person}{Yiwei Lyu},
  \bibinfo{person}{Xiang Fan}, \bibinfo{person}{Shengtong Mo},
  \bibinfo{person}{Dani Yogatama}, \bibinfo{person}{Louis-Philippe Morency},
  {and} \bibinfo{person}{Ruslan Salakhutdinov}.}
  \bibinfo{year}{2022}\natexlab{b}.
\newblock \showarticletitle{Highmmt: Towards modality and task generalization
  for high-modality representation learning}.
\newblock \bibinfo{journal}{\emph{arXiv preprint arXiv:2203.01311}}
  (\bibinfo{year}{2022}).
\newblock


\bibitem[Liang et~al\mbox{.}({[n.\,d.]})]%
        {lianghighmmt}
\bibfield{author}{\bibinfo{person}{Paul~Pu Liang}, \bibinfo{person}{Yiwei Lyu},
  \bibinfo{person}{Xiang Fan}, \bibinfo{person}{Jeffrey Tsaw},
  \bibinfo{person}{Yudong Liu}, \bibinfo{person}{Shentong Mo},
  \bibinfo{person}{Dani Yogatama}, \bibinfo{person}{Louis-Philippe Morency},
  {and} \bibinfo{person}{Ruslan Salakhutdinov}.}
  \bibinfo{year}{[n.\,d.]}\natexlab{}.
\newblock \showarticletitle{HighMMT: Quantifying Modality \& Interaction
  Heterogeneity for High-Modality Representation Learning}.
\newblock  (\bibinfo{year}{[n.\,d.]}).
\newblock


\bibitem[Liang et~al\mbox{.}(2021)]%
        {liang2021multibench}
\bibfield{author}{\bibinfo{person}{Paul~Pu Liang}, \bibinfo{person}{Yiwei Lyu},
  \bibinfo{person}{Xiang Fan}, \bibinfo{person}{Zetian Wu},
  \bibinfo{person}{Yun Cheng}, \bibinfo{person}{Jason Wu},
  \bibinfo{person}{Leslie Chen}, \bibinfo{person}{Peter Wu},
  \bibinfo{person}{Michelle~A Lee}, \bibinfo{person}{Yuke Zhu},
  {et~al\mbox{.}}} \bibinfo{year}{2021}\natexlab{}.
\newblock \showarticletitle{Multibench: Multiscale benchmarks for multimodal
  representation learning}.
\newblock \bibinfo{journal}{\emph{arXiv preprint arXiv:2107.07502}}
  (\bibinfo{year}{2021}).
\newblock


\bibitem[Liang et~al\mbox{.}(2022c)]%
        {liang2022foundations}
\bibfield{author}{\bibinfo{person}{Paul~Pu Liang}, \bibinfo{person}{Amir
  Zadeh}, {and} \bibinfo{person}{Louis-Philippe Morency}.}
  \bibinfo{year}{2022}\natexlab{c}.
\newblock \showarticletitle{Foundations and recent trends in multimodal machine
  learning: Principles, challenges, and open questions}.
\newblock \bibinfo{journal}{\emph{arXiv preprint arXiv:2209.03430}}
  (\bibinfo{year}{2022}).
\newblock


\bibitem[Lin et~al\mbox{.}(2023)]%
        {lin2023generating}
\bibfield{author}{\bibinfo{person}{Zhen Lin}, \bibinfo{person}{Shubhendu
  Trivedi}, {and} \bibinfo{person}{Jimeng Sun}.}
  \bibinfo{year}{2023}\natexlab{}.
\newblock \showarticletitle{Generating with Confidence: Uncertainty
  Quantification for Black-box Large Language Models}.
\newblock \bibinfo{journal}{\emph{arXiv preprint arXiv:2305.19187}}
  (\bibinfo{year}{2023}).
\newblock


\bibitem[Liu et~al\mbox{.}(2023a)]%
        {liu2023audioldm}
\bibfield{author}{\bibinfo{person}{Haohe Liu}, \bibinfo{person}{Zehua Chen},
  \bibinfo{person}{Yi Yuan}, \bibinfo{person}{Xinhao Mei},
  \bibinfo{person}{Xubo Liu}, \bibinfo{person}{Danilo Mandic},
  \bibinfo{person}{Wenwu Wang}, {and} \bibinfo{person}{Mark~D Plumbley}.}
  \bibinfo{year}{2023}\natexlab{a}.
\newblock \showarticletitle{Audioldm: Text-to-audio generation with latent
  diffusion models}.
\newblock \bibinfo{journal}{\emph{arXiv preprint arXiv:2301.12503}}
  (\bibinfo{year}{2023}).
\newblock


\bibitem[Liu et~al\mbox{.}(2023b)]%
        {liu2023improved}
\bibfield{author}{\bibinfo{person}{Haotian Liu}, \bibinfo{person}{Chunyuan Li},
  \bibinfo{person}{Yuheng Li}, {and} \bibinfo{person}{Yong~Jae Lee}.}
  \bibinfo{year}{2023}\natexlab{b}.
\newblock \showarticletitle{Improved baselines with visual instruction tuning}.
\newblock \bibinfo{journal}{\emph{arXiv preprint arXiv:2310.03744}}
  (\bibinfo{year}{2023}).
\newblock


\bibitem[Liu et~al\mbox{.}(2023c)]%
        {liu2023visual}
\bibfield{author}{\bibinfo{person}{Haotian Liu}, \bibinfo{person}{Chunyuan Li},
  \bibinfo{person}{Qingyang Wu}, {and} \bibinfo{person}{Yong~Jae Lee}.}
  \bibinfo{year}{2023}\natexlab{c}.
\newblock \showarticletitle{Visual instruction tuning}.
\newblock \bibinfo{journal}{\emph{arXiv preprint arXiv:2304.08485}}
  (\bibinfo{year}{2023}).
\newblock


\bibitem[Liu et~al\mbox{.}(2023d)]%
        {liu2023towards}
\bibfield{author}{\bibinfo{person}{Jiawei Liu}, \bibinfo{person}{Cheng Yang},
  \bibinfo{person}{Zhiyuan Lu}, \bibinfo{person}{Junze Chen},
  \bibinfo{person}{Yibo Li}, \bibinfo{person}{Mengmei Zhang},
  \bibinfo{person}{Ting Bai}, \bibinfo{person}{Yuan Fang},
  \bibinfo{person}{Lichao Sun}, \bibinfo{person}{Philip~S Yu}, {et~al\mbox{.}}}
  \bibinfo{year}{2023}\natexlab{d}.
\newblock \showarticletitle{Towards Graph Foundation Models: A Survey and
  Beyond}.
\newblock \bibinfo{journal}{\emph{arXiv preprint arXiv:2310.11829}}
  (\bibinfo{year}{2023}).
\newblock


\bibitem[Liu et~al\mbox{.}(2021)]%
        {liu2021swin}
\bibfield{author}{\bibinfo{person}{Ze Liu}, \bibinfo{person}{Yutong Lin},
  \bibinfo{person}{Yue Cao}, \bibinfo{person}{Han Hu}, \bibinfo{person}{Yixuan
  Wei}, \bibinfo{person}{Zheng Zhang}, \bibinfo{person}{Stephen Lin}, {and}
  \bibinfo{person}{Baining Guo}.} \bibinfo{year}{2021}\natexlab{}.
\newblock \showarticletitle{Swin transformer: Hierarchical vision transformer
  using shifted windows}. In \bibinfo{booktitle}{\emph{Proceedings of the
  IEEE/CVF international conference on computer vision}}.
  \bibinfo{pages}{10012--10022}.
\newblock


\bibitem[Lu et~al\mbox{.}(2022)]%
        {lu2022unified}
\bibfield{author}{\bibinfo{person}{Jiasen Lu}, \bibinfo{person}{Christopher
  Clark}, \bibinfo{person}{Rowan Zellers}, \bibinfo{person}{Roozbeh Mottaghi},
  {and} \bibinfo{person}{Aniruddha Kembhavi}.} \bibinfo{year}{2022}\natexlab{}.
\newblock \showarticletitle{Unified-io: A unified model for vision, language,
  and multi-modal tasks}.
\newblock \bibinfo{journal}{\emph{arXiv preprint arXiv:2206.08916}}
  (\bibinfo{year}{2022}).
\newblock


\bibitem[Lu et~al\mbox{.}(2023)]%
        {lu2023empirical}
\bibfield{author}{\bibinfo{person}{Yadong Lu}, \bibinfo{person}{Chunyuan Li},
  \bibinfo{person}{Haotian Liu}, \bibinfo{person}{Jianwei Yang},
  \bibinfo{person}{Jianfeng Gao}, {and} \bibinfo{person}{Yelong Shen}.}
  \bibinfo{year}{2023}\natexlab{}.
\newblock \showarticletitle{An empirical study of scaling instruct-tuned large
  multimodal models}.
\newblock \bibinfo{journal}{\emph{arXiv preprint arXiv:2309.09958}}
  (\bibinfo{year}{2023}).
\newblock


\bibitem[Lyu et~al\mbox{.}(2023)]%
        {lyu2023macaw}
\bibfield{author}{\bibinfo{person}{Chenyang Lyu}, \bibinfo{person}{Minghao Wu},
  \bibinfo{person}{Longyue Wang}, \bibinfo{person}{Xinting Huang},
  \bibinfo{person}{Bingshuai Liu}, \bibinfo{person}{Zefeng Du},
  \bibinfo{person}{Shuming Shi}, {and} \bibinfo{person}{Zhaopeng Tu}.}
  \bibinfo{year}{2023}\natexlab{}.
\newblock \showarticletitle{Macaw-LLM: Multi-Modal Language Modeling with
  Image, Audio, Video, and Text Integration}.
\newblock \bibinfo{journal}{\emph{arXiv preprint arXiv:2306.09093}}
  (\bibinfo{year}{2023}).
\newblock


\bibitem[Matsuo et~al\mbox{.}(2022)]%
        {matsuo2022deep}
\bibfield{author}{\bibinfo{person}{Yutaka Matsuo}, \bibinfo{person}{Yann
  LeCun}, \bibinfo{person}{Maneesh Sahani}, \bibinfo{person}{Doina Precup},
  \bibinfo{person}{David Silver}, \bibinfo{person}{Masashi Sugiyama},
  \bibinfo{person}{Eiji Uchibe}, {and} \bibinfo{person}{Jun Morimoto}.}
  \bibinfo{year}{2022}\natexlab{}.
\newblock \showarticletitle{Deep learning, reinforcement learning, and world
  models}.
\newblock \bibinfo{journal}{\emph{Neural Networks}} (\bibinfo{year}{2022}).
\newblock


\bibitem[Muhammad et~al\mbox{.}(2021)]%
        {muhammad2021comprehensive}
\bibfield{author}{\bibinfo{person}{Ghulam Muhammad}, \bibinfo{person}{Fatima
  Alshehri}, \bibinfo{person}{Fakhri Karray}, \bibinfo{person}{Abdulmotaleb
  El~Saddik}, \bibinfo{person}{Mansour Alsulaiman}, {and}
  \bibinfo{person}{Tiago~H Falk}.} \bibinfo{year}{2021}\natexlab{}.
\newblock \showarticletitle{A comprehensive survey on multimodal medical
  signals fusion for smart healthcare systems}.
\newblock \bibinfo{journal}{\emph{Information Fusion}}  \bibinfo{volume}{76}
  (\bibinfo{year}{2021}), \bibinfo{pages}{355--375}.
\newblock


\bibitem[Munikoti et~al\mbox{.}(2023)]%
        {munikoti2023atlantic}
\bibfield{author}{\bibinfo{person}{Sai Munikoti}, \bibinfo{person}{Anurag
  Acharya}, \bibinfo{person}{Sridevi Wagle}, {and} \bibinfo{person}{Sameera
  Horawalavithana}.} \bibinfo{year}{2023}\natexlab{}.
\newblock \showarticletitle{ATLANTIC: Structure-Aware Retrieval-Augmented
  Language Model for Interdisciplinary Science}.
\newblock \bibinfo{journal}{\emph{arXiv preprint arXiv:2311.12289}}
  (\bibinfo{year}{2023}).
\newblock


\bibitem[Peng et~al\mbox{.}(2020)]%
        {peng2020self}
\bibfield{author}{\bibinfo{person}{Zhen Peng}, \bibinfo{person}{Yixiang Dong},
  \bibinfo{person}{Minnan Luo}, \bibinfo{person}{Xiao-Ming Wu}, {and}
  \bibinfo{person}{Qinghua Zheng}.} \bibinfo{year}{2020}\natexlab{}.
\newblock \showarticletitle{Self-supervised graph representation learning via
  global context prediction}.
\newblock \bibinfo{journal}{\emph{arXiv preprint arXiv:2003.01604}}
  (\bibinfo{year}{2020}).
\newblock


\bibitem[Radford et~al\mbox{.}(2021)]%
        {radford2021learning}
\bibfield{author}{\bibinfo{person}{Alec Radford}, \bibinfo{person}{Jong~Wook
  Kim}, \bibinfo{person}{Chris Hallacy}, \bibinfo{person}{Aditya Ramesh},
  \bibinfo{person}{Gabriel Goh}, \bibinfo{person}{Sandhini Agarwal},
  \bibinfo{person}{Girish Sastry}, \bibinfo{person}{Amanda Askell},
  \bibinfo{person}{Pamela Mishkin}, \bibinfo{person}{Jack Clark},
  {et~al\mbox{.}}} \bibinfo{year}{2021}\natexlab{}.
\newblock \showarticletitle{Learning transferable visual models from natural
  language supervision}. In \bibinfo{booktitle}{\emph{International conference
  on machine learning}}. PMLR, \bibinfo{pages}{8748--8763}.
\newblock


\bibitem[Radford et~al\mbox{.}(2023)]%
        {radford2023robust}
\bibfield{author}{\bibinfo{person}{Alec Radford}, \bibinfo{person}{Jong~Wook
  Kim}, \bibinfo{person}{Tao Xu}, \bibinfo{person}{Greg Brockman},
  \bibinfo{person}{Christine McLeavey}, {and} \bibinfo{person}{Ilya
  Sutskever}.} \bibinfo{year}{2023}\natexlab{}.
\newblock \showarticletitle{Robust speech recognition via large-scale weak
  supervision}. In \bibinfo{booktitle}{\emph{International Conference on
  Machine Learning}}. PMLR, \bibinfo{pages}{28492--28518}.
\newblock


\bibitem[Raffel et~al\mbox{.}(2020)]%
        {raffel2020exploring}
\bibfield{author}{\bibinfo{person}{Colin Raffel}, \bibinfo{person}{Noam
  Shazeer}, \bibinfo{person}{Adam Roberts}, \bibinfo{person}{Katherine Lee},
  \bibinfo{person}{Sharan Narang}, \bibinfo{person}{Michael Matena},
  \bibinfo{person}{Yanqi Zhou}, \bibinfo{person}{Wei Li}, {and}
  \bibinfo{person}{Peter~J Liu}.} \bibinfo{year}{2020}\natexlab{}.
\newblock \showarticletitle{Exploring the limits of transfer learning with a
  unified text-to-text transformer}.
\newblock \bibinfo{journal}{\emph{The Journal of Machine Learning Research}}
  \bibinfo{volume}{21}, \bibinfo{number}{1} (\bibinfo{year}{2020}),
  \bibinfo{pages}{5485--5551}.
\newblock


\bibitem[Rajpurkar et~al\mbox{.}(2018)]%
        {rajpurkar2018know}
\bibfield{author}{\bibinfo{person}{Pranav Rajpurkar}, \bibinfo{person}{Robin
  Jia}, {and} \bibinfo{person}{Percy Liang}.} \bibinfo{year}{2018}\natexlab{}.
\newblock \showarticletitle{Know What You Don't Know: Unanswerable Questions
  for {SQuAD}}. In \bibinfo{booktitle}{\emph{Proceedings of the 56th Annual
  Meeting of the Association for Computational Linguistics (Volume 2: Short
  Papers)}}. Association for Computational Linguistics.
\newblock


\bibitem[Ramesh et~al\mbox{.}(2021)]%
        {ramesh2021zero}
\bibfield{author}{\bibinfo{person}{Aditya Ramesh}, \bibinfo{person}{Mikhail
  Pavlov}, \bibinfo{person}{Gabriel Goh}, \bibinfo{person}{Scott Gray},
  \bibinfo{person}{Chelsea Voss}, \bibinfo{person}{Alec Radford},
  \bibinfo{person}{Mark Chen}, {and} \bibinfo{person}{Ilya Sutskever}.}
  \bibinfo{year}{2021}\natexlab{}.
\newblock \showarticletitle{Zero-shot text-to-image generation}. In
  \bibinfo{booktitle}{\emph{International Conference on Machine Learning}}.
  PMLR, \bibinfo{pages}{8821--8831}.
\newblock


\bibitem[Reed et~al\mbox{.}(2022)]%
        {reed2022generalist}
\bibfield{author}{\bibinfo{person}{Scott Reed}, \bibinfo{person}{Konrad Zolna},
  \bibinfo{person}{Emilio Parisotto}, \bibinfo{person}{Sergio~Gomez
  Colmenarejo}, \bibinfo{person}{Alexander Novikov}, \bibinfo{person}{Gabriel
  Barth-Maron}, \bibinfo{person}{Mai Gimenez}, \bibinfo{person}{Yury Sulsky},
  \bibinfo{person}{Jackie Kay}, \bibinfo{person}{Jost~Tobias Springenberg},
  {et~al\mbox{.}}} \bibinfo{year}{2022}\natexlab{}.
\newblock \showarticletitle{A generalist agent}.
\newblock \bibinfo{journal}{\emph{arXiv preprint arXiv:2205.06175}}
  (\bibinfo{year}{2022}).
\newblock


\bibitem[Ren et~al\mbox{.}(2023)]%
        {ren2023robots}
\bibfield{author}{\bibinfo{person}{Allen~Z Ren}, \bibinfo{person}{Anushri
  Dixit}, \bibinfo{person}{Alexandra Bodrova}, \bibinfo{person}{Sumeet Singh},
  \bibinfo{person}{Stephen Tu}, \bibinfo{person}{Noah Brown},
  \bibinfo{person}{Peng Xu}, \bibinfo{person}{Leila Takayama},
  \bibinfo{person}{Fei Xia}, \bibinfo{person}{Jake Varley}, {et~al\mbox{.}}}
  \bibinfo{year}{2023}\natexlab{}.
\newblock \showarticletitle{Robots that ask for help: Uncertainty alignment for
  large language model planners}.
\newblock \bibinfo{journal}{\emph{arXiv preprint arXiv:2307.01928}}
  (\bibinfo{year}{2023}).
\newblock


\bibitem[Rombach et~al\mbox{.}(2021)]%
        {rombach2021high}
\bibfield{author}{\bibinfo{person}{Robin Rombach}, \bibinfo{person}{Andreas
  Blattmann}, \bibinfo{person}{Dominik Lorenz}, \bibinfo{person}{Patrick
  Esser}, {and} \bibinfo{person}{Bj{\"o}rn Ommer}.}
  \bibinfo{year}{2021}\natexlab{}.
\newblock \showarticletitle{High-resolution image synthesis with latent
  diffusion models. 2022 IEEE}. In \bibinfo{booktitle}{\emph{CVF Conference on
  Computer Vision and Pattern Recognition (CVPR)}}.
  \bibinfo{pages}{10674--10685}.
\newblock


\bibitem[Rong et~al\mbox{.}(2020)]%
        {rong2020self}
\bibfield{author}{\bibinfo{person}{Yu Rong}, \bibinfo{person}{Yatao Bian},
  \bibinfo{person}{Tingyang Xu}, \bibinfo{person}{Weiyang Xie},
  \bibinfo{person}{Ying Wei}, \bibinfo{person}{Wenbing Huang}, {and}
  \bibinfo{person}{Junzhou Huang}.} \bibinfo{year}{2020}\natexlab{}.
\newblock \showarticletitle{Self-supervised graph transformer on large-scale
  molecular data}.
\newblock \bibinfo{journal}{\emph{Advances in Neural Information Processing
  Systems}}  \bibinfo{volume}{33} (\bibinfo{year}{2020}),
  \bibinfo{pages}{12559--12571}.
\newblock


\bibitem[Sanh et~al\mbox{.}(2021)]%
        {sanh2021multitask}
\bibfield{author}{\bibinfo{person}{Victor Sanh}, \bibinfo{person}{Albert
  Webson}, \bibinfo{person}{Colin Raffel}, \bibinfo{person}{Stephen~H Bach},
  \bibinfo{person}{Lintang Sutawika}, \bibinfo{person}{Zaid Alyafeai},
  \bibinfo{person}{Antoine Chaffin}, \bibinfo{person}{Arnaud Stiegler},
  \bibinfo{person}{Teven~Le Scao}, \bibinfo{person}{Arun Raja},
  {et~al\mbox{.}}} \bibinfo{year}{2021}\natexlab{}.
\newblock \showarticletitle{Multitask prompted training enables zero-shot task
  generalization}.
\newblock \bibinfo{journal}{\emph{arXiv preprint arXiv:2110.08207}}
  (\bibinfo{year}{2021}).
\newblock


\bibitem[Singh et~al\mbox{.}(2022)]%
        {singh2022flava}
\bibfield{author}{\bibinfo{person}{Amanpreet Singh}, \bibinfo{person}{Ronghang
  Hu}, \bibinfo{person}{Vedanuj Goswami}, \bibinfo{person}{Guillaume Couairon},
  \bibinfo{person}{Wojciech Galuba}, \bibinfo{person}{Marcus Rohrbach}, {and}
  \bibinfo{person}{Douwe Kiela}.} \bibinfo{year}{2022}\natexlab{}.
\newblock \showarticletitle{Flava: A foundational language and vision alignment
  model}. In \bibinfo{booktitle}{\emph{Proceedings of the IEEE/CVF Conference
  on Computer Vision and Pattern Recognition}}. \bibinfo{pages}{15638--15650}.
\newblock


\bibitem[Stepin et~al\mbox{.}(2021)]%
        {stepin2021survey}
\bibfield{author}{\bibinfo{person}{Ilia Stepin}, \bibinfo{person}{Jose~M
  Alonso}, \bibinfo{person}{Alejandro Catala}, {and}
  \bibinfo{person}{Mart{\'\i}n Pereira-Fari{\~n}a}.}
  \bibinfo{year}{2021}\natexlab{}.
\newblock \showarticletitle{A survey of contrastive and counterfactual
  explanation generation methods for explainable artificial intelligence}.
\newblock \bibinfo{journal}{\emph{IEEE Access}}  \bibinfo{volume}{9}
  (\bibinfo{year}{2021}), \bibinfo{pages}{11974--12001}.
\newblock


\bibitem[Touvron et~al\mbox{.}(2023)]%
        {touvron2023llama}
\bibfield{author}{\bibinfo{person}{Hugo Touvron}, \bibinfo{person}{Thibaut
  Lavril}, \bibinfo{person}{Gautier Izacard}, \bibinfo{person}{Xavier
  Martinet}, \bibinfo{person}{Marie-Anne Lachaux},
  \bibinfo{person}{Timoth{\'e}e Lacroix}, \bibinfo{person}{Baptiste
  Rozi{\`e}re}, \bibinfo{person}{Naman Goyal}, \bibinfo{person}{Eric Hambro},
  \bibinfo{person}{Faisal Azhar}, {et~al\mbox{.}}}
  \bibinfo{year}{2023}\natexlab{}.
\newblock \showarticletitle{Llama: Open and efficient foundation language
  models}.
\newblock \bibinfo{journal}{\emph{arXiv preprint arXiv:2302.13971}}
  (\bibinfo{year}{2023}).
\newblock


\bibitem[Van Den~Oord et~al\mbox{.}(2017)]%
        {van2017neural}
\bibfield{author}{\bibinfo{person}{Aaron Van Den~Oord}, \bibinfo{person}{Oriol
  Vinyals}, {et~al\mbox{.}}} \bibinfo{year}{2017}\natexlab{}.
\newblock \showarticletitle{Neural discrete representation learning}.
\newblock \bibinfo{journal}{\emph{Advances in neural information processing
  systems}}  \bibinfo{volume}{30} (\bibinfo{year}{2017}).
\newblock


\bibitem[Vaswani et~al\mbox{.}(2017)]%
        {vaswani2017attention}
\bibfield{author}{\bibinfo{person}{Ashish Vaswani}, \bibinfo{person}{Noam
  Shazeer}, \bibinfo{person}{Niki Parmar}, \bibinfo{person}{Jakob Uszkoreit},
  \bibinfo{person}{Llion Jones}, \bibinfo{person}{Aidan~N Gomez},
  \bibinfo{person}{{\L}ukasz Kaiser}, {and} \bibinfo{person}{Illia
  Polosukhin}.} \bibinfo{year}{2017}\natexlab{}.
\newblock \showarticletitle{Attention is all you need}.
\newblock \bibinfo{journal}{\emph{Advances in neural information processing
  systems}}  \bibinfo{volume}{30} (\bibinfo{year}{2017}).
\newblock


\bibitem[Verma et~al\mbox{.}(2022)]%
        {verma2022attention}
\bibfield{author}{\bibinfo{person}{Shailza Verma}, \bibinfo{person}{Neelansha
  Srivastava}, \bibinfo{person}{Piyush Rai}, {and} \bibinfo{person}{Vasudha
  Ahuja}.} \bibinfo{year}{2022}\natexlab{}.
\newblock \showarticletitle{Attention Mechanism-based Transformer Model for
  Predicting Parkinson’s Disease Progression}.
\newblock \bibinfo{journal}{\emph{Journal of Ambient Intelligence and Humanized
  Computing}}  \bibinfo{volume}{13} (\bibinfo{year}{2022}),
  \bibinfo{pages}{6155--6164}.
\newblock


\bibitem[Wagle et~al\mbox{.}(2023)]%
        {wagle2023empirical}
\bibfield{author}{\bibinfo{person}{Sridevi Wagle}, \bibinfo{person}{Sai
  Munikoti}, \bibinfo{person}{Anurag Acharya}, \bibinfo{person}{Sara Smith},
  {and} \bibinfo{person}{Sameera Horawalavithana}.}
  \bibinfo{year}{2023}\natexlab{}.
\newblock \showarticletitle{Empirical evaluation of Uncertainty Quantification
  in Retrieval-Augmented Language Models for Science}.
\newblock \bibinfo{journal}{\emph{arXiv preprint arXiv:2311.09358}}
  (\bibinfo{year}{2023}).
\newblock


\bibitem[Wang et~al\mbox{.}(2023b)]%
        {wang2023voyager}
\bibfield{author}{\bibinfo{person}{Guanzhi Wang}, \bibinfo{person}{Yuqi Xie},
  \bibinfo{person}{Yunfan Jiang}, \bibinfo{person}{Ajay Mandlekar},
  \bibinfo{person}{Chaowei Xiao}, \bibinfo{person}{Yuke Zhu},
  \bibinfo{person}{Linxi Fan}, {and} \bibinfo{person}{Anima Anandkumar}.}
  \bibinfo{year}{2023}\natexlab{b}.
\newblock \showarticletitle{Voyager: An open-ended embodied agent with large
  language models}.
\newblock \bibinfo{journal}{\emph{arXiv preprint arXiv:2305.16291}}
  (\bibinfo{year}{2023}).
\newblock


\bibitem[Wang et~al\mbox{.}(2022a)]%
        {wang2022omnivl}
\bibfield{author}{\bibinfo{person}{Junke Wang}, \bibinfo{person}{Dongdong
  Chen}, \bibinfo{person}{Zuxuan Wu}, \bibinfo{person}{Chong Luo},
  \bibinfo{person}{Luowei Zhou}, \bibinfo{person}{Yucheng Zhao},
  \bibinfo{person}{Yujia Xie}, \bibinfo{person}{Ce Liu},
  \bibinfo{person}{Yu-Gang Jiang}, {and} \bibinfo{person}{Lu Yuan}.}
  \bibinfo{year}{2022}\natexlab{a}.
\newblock \showarticletitle{Omnivl: One foundation model for image-language and
  video-language tasks}.
\newblock \bibinfo{journal}{\emph{arXiv preprint arXiv:2209.07526}}
  (\bibinfo{year}{2022}).
\newblock


\bibitem[Wang et~al\mbox{.}(2022d)]%
        {wang2022ofa}
\bibfield{author}{\bibinfo{person}{Peng Wang}, \bibinfo{person}{An Yang},
  \bibinfo{person}{Rui Men}, \bibinfo{person}{Junyang Lin},
  \bibinfo{person}{Shuai Bai}, \bibinfo{person}{Zhikang Li},
  \bibinfo{person}{Jianxin Ma}, \bibinfo{person}{Chang Zhou},
  \bibinfo{person}{Jingren Zhou}, {and} \bibinfo{person}{Hongxia Yang}.}
  \bibinfo{year}{2022}\natexlab{d}.
\newblock \showarticletitle{Ofa: Unifying architectures, tasks, and modalities
  through a simple sequence-to-sequence learning framework}. In
  \bibinfo{booktitle}{\emph{International Conference on Machine Learning}}.
  PMLR, \bibinfo{pages}{23318--23340}.
\newblock


\bibitem[Wang et~al\mbox{.}(2022b)]%
        {wang2022internimage}
\bibfield{author}{\bibinfo{person}{Wenhai Wang}, \bibinfo{person}{Jifeng Dai},
  \bibinfo{person}{Zhe Chen}, \bibinfo{person}{Zhenhang Huang},
  \bibinfo{person}{Zhiqi Li}, \bibinfo{person}{Xizhou Zhu},
  \bibinfo{person}{Xiaowei Hu}, \bibinfo{person}{Tong Lu},
  \bibinfo{person}{Lewei Lu}, \bibinfo{person}{Hongsheng Li}, {et~al\mbox{.}}}
  \bibinfo{year}{2022}\natexlab{b}.
\newblock \showarticletitle{Internimage: Exploring large-scale vision
  foundation models with deformable convolutions}.
\newblock \bibinfo{journal}{\emph{arXiv preprint arXiv:2211.05778}}
  (\bibinfo{year}{2022}).
\newblock


\bibitem[Wang et~al\mbox{.}(2023a)]%
        {wang2023internimage}
\bibfield{author}{\bibinfo{person}{Wenhai Wang}, \bibinfo{person}{Jifeng Dai},
  \bibinfo{person}{Zhe Chen}, \bibinfo{person}{Zhenhang Huang},
  \bibinfo{person}{Zhiqi Li}, \bibinfo{person}{Xizhou Zhu},
  \bibinfo{person}{Xiaowei Hu}, \bibinfo{person}{Tong Lu},
  \bibinfo{person}{Lewei Lu}, \bibinfo{person}{Hongsheng Li}, {et~al\mbox{.}}}
  \bibinfo{year}{2023}\natexlab{a}.
\newblock \showarticletitle{Internimage: Exploring large-scale vision
  foundation models with deformable convolutions}. In
  \bibinfo{booktitle}{\emph{Proceedings of the IEEE/CVF Conference on Computer
  Vision and Pattern Recognition}}. \bibinfo{pages}{14408--14419}.
\newblock


\bibitem[Wang et~al\mbox{.}(2022c)]%
        {wang2022internvideo}
\bibfield{author}{\bibinfo{person}{Yi Wang}, \bibinfo{person}{Kunchang Li},
  \bibinfo{person}{Yizhuo Li}, \bibinfo{person}{Yinan He},
  \bibinfo{person}{Bingkun Huang}, \bibinfo{person}{Zhiyu Zhao},
  \bibinfo{person}{Hongjie Zhang}, \bibinfo{person}{Jilan Xu},
  \bibinfo{person}{Yi Liu}, \bibinfo{person}{Zun Wang}, {et~al\mbox{.}}}
  \bibinfo{year}{2022}\natexlab{c}.
\newblock \showarticletitle{InternVideo: General Video Foundation Models via
  Generative and Discriminative Learning}.
\newblock \bibinfo{journal}{\emph{arXiv preprint arXiv:2212.03191}}
  (\bibinfo{year}{2022}).
\newblock


\bibitem[Wang et~al\mbox{.}(2021)]%
        {wang2021simvlm}
\bibfield{author}{\bibinfo{person}{Zirui Wang}, \bibinfo{person}{Jiahui Yu},
  \bibinfo{person}{Adams~Wei Yu}, \bibinfo{person}{Zihang Dai},
  \bibinfo{person}{Yulia Tsvetkov}, {and} \bibinfo{person}{Yuan Cao}.}
  \bibinfo{year}{2021}\natexlab{}.
\newblock \showarticletitle{Simvlm: Simple visual language model pretraining
  with weak supervision}.
\newblock \bibinfo{journal}{\emph{arXiv preprint arXiv:2108.10904}}
  (\bibinfo{year}{2021}).
\newblock


\bibitem[Weinbach et~al\mbox{.}(2022)]%
        {weinbach2022m}
\bibfield{author}{\bibinfo{person}{Samuel Weinbach}, \bibinfo{person}{Marco
  Bellagente}, \bibinfo{person}{Constantin Eichenberg}, \bibinfo{person}{Andrew
  Dai}, \bibinfo{person}{Robert Baldock}, \bibinfo{person}{Souradeep Nanda},
  \bibinfo{person}{Bj{\"o}rn Deiseroth}, \bibinfo{person}{Koen Oostermeijer},
  \bibinfo{person}{Hannah Teufel}, {and} \bibinfo{person}{Andres~Felipe
  Cruz-Salinas}.} \bibinfo{year}{2022}\natexlab{}.
\newblock \showarticletitle{M-VADER: A Model for Diffusion with Multimodal
  Context}.
\newblock \bibinfo{journal}{\emph{arXiv preprint arXiv:2212.02936}}
  (\bibinfo{year}{2022}).
\newblock


\bibitem[Wu et~al\mbox{.}(2020)]%
        {wu2020deep}
\bibfield{author}{\bibinfo{person}{Neo Wu}, \bibinfo{person}{Bradley Green},
  \bibinfo{person}{Xue Ben}, {and} \bibinfo{person}{Shawn O'Banion}.}
  \bibinfo{year}{2020}\natexlab{}.
\newblock \showarticletitle{Deep transformer models for time series
  forecasting: The influenza prevalence case}.
\newblock \bibinfo{journal}{\emph{arXiv preprint arXiv:2001.08317}}
  (\bibinfo{year}{2020}).
\newblock


\bibitem[Wu et~al\mbox{.}(2023)]%
        {wu2023next}
\bibfield{author}{\bibinfo{person}{Shengqiong Wu}, \bibinfo{person}{Hao Fei},
  \bibinfo{person}{Leigang Qu}, \bibinfo{person}{Wei Ji}, {and}
  \bibinfo{person}{Tat-Seng Chua}.} \bibinfo{year}{2023}\natexlab{}.
\newblock \showarticletitle{NExT-GPT: Any-to-Any Multimodal LLM}.
\newblock \bibinfo{journal}{\emph{arXiv preprint arXiv:2309.05519}}
  (\bibinfo{year}{2023}).
\newblock


\bibitem[Xiao et~al\mbox{.}(2022)]%
        {xiao2022uncertainty}
\bibfield{author}{\bibinfo{person}{Yuxin Xiao}, \bibinfo{person}{Paul~Pu
  Liang}, \bibinfo{person}{Umang Bhatt}, \bibinfo{person}{Willie Neiswanger},
  \bibinfo{person}{Ruslan Salakhutdinov}, {and} \bibinfo{person}{Louis-Philippe
  Morency}.} \bibinfo{year}{2022}\natexlab{}.
\newblock \showarticletitle{Uncertainty Quantification with Pre-trained
  Language Models: A Large-Scale Empirical Analysis}.
\newblock \bibinfo{journal}{\emph{arXiv preprint arXiv:2210.04714}}
  (\bibinfo{year}{2022}).
\newblock


\bibitem[Xu et~al\mbox{.}(2023)]%
        {xu2023mplug}
\bibfield{author}{\bibinfo{person}{Haiyang Xu}, \bibinfo{person}{Qinghao Ye},
  \bibinfo{person}{Ming Yan}, \bibinfo{person}{Yaya Shi},
  \bibinfo{person}{Jiabo Ye}, \bibinfo{person}{Yuanhong Xu},
  \bibinfo{person}{Chenliang Li}, \bibinfo{person}{Bin Bi}, \bibinfo{person}{Qi
  Qian}, \bibinfo{person}{Wei Wang}, {et~al\mbox{.}}}
  \bibinfo{year}{2023}\natexlab{}.
\newblock \showarticletitle{mPLUG-2: A Modularized Multi-modal Foundation Model
  Across Text, Image and Video}.
\newblock \bibinfo{journal}{\emph{arXiv preprint arXiv:2302.00402}}
  (\bibinfo{year}{2023}).
\newblock


\bibitem[Xue et~al\mbox{.}(2020)]%
        {xue2020mt5}
\bibfield{author}{\bibinfo{person}{Linting Xue}, \bibinfo{person}{Noah
  Constant}, \bibinfo{person}{Adam Roberts}, \bibinfo{person}{Mihir Kale},
  \bibinfo{person}{Rami Al-Rfou}, \bibinfo{person}{Aditya Siddhant},
  \bibinfo{person}{Aditya Barua}, {and} \bibinfo{person}{Colin Raffel}.}
  \bibinfo{year}{2020}\natexlab{}.
\newblock \showarticletitle{mT5: A massively multilingual pre-trained
  text-to-text transformer}.
\newblock \bibinfo{journal}{\emph{arXiv preprint arXiv:2010.11934}}
  (\bibinfo{year}{2020}).
\newblock


\bibitem[Yang et~al\mbox{.}(2022)]%
        {yang2022prompt}
\bibfield{author}{\bibinfo{person}{Hao Yang}, \bibinfo{person}{Junyang Lin},
  \bibinfo{person}{An Yang}, \bibinfo{person}{Peng Wang},
  \bibinfo{person}{Chang Zhou}, {and} \bibinfo{person}{Hongxia Yang}.}
  \bibinfo{year}{2022}\natexlab{}.
\newblock \showarticletitle{Prompt tuning for generative multimodal pretrained
  models}.
\newblock \bibinfo{journal}{\emph{arXiv preprint arXiv:2208.02532}}
  (\bibinfo{year}{2022}).
\newblock


\bibitem[Yang et~al\mbox{.}(2023)]%
        {yang2023foundation}
\bibfield{author}{\bibinfo{person}{Sherry Yang}, \bibinfo{person}{Ofir Nachum},
  \bibinfo{person}{Yilun Du}, \bibinfo{person}{Jason Wei},
  \bibinfo{person}{Pieter Abbeel}, {and} \bibinfo{person}{Dale Schuurmans}.}
  \bibinfo{year}{2023}\natexlab{}.
\newblock \showarticletitle{Foundation Models for Decision Making: Problems,
  Methods, and Opportunities}.
\newblock \bibinfo{journal}{\emph{arXiv preprint arXiv:2303.04129}}
  (\bibinfo{year}{2023}).
\newblock


\bibitem[Yao et~al\mbox{.}(2022)]%
        {yao2022pevl}
\bibfield{author}{\bibinfo{person}{Yuan Yao}, \bibinfo{person}{Qianyu Chen},
  \bibinfo{person}{Ao Zhang}, \bibinfo{person}{Wei Ji},
  \bibinfo{person}{Zhiyuan Liu}, \bibinfo{person}{Tat-Seng Chua}, {and}
  \bibinfo{person}{Maosong Sun}.} \bibinfo{year}{2022}\natexlab{}.
\newblock \showarticletitle{PEVL: Position-enhanced pre-training and prompt
  tuning for vision-language models}.
\newblock \bibinfo{journal}{\emph{arXiv preprint arXiv:2205.11169}}
  (\bibinfo{year}{2022}).
\newblock


\bibitem[Yasunaga et~al\mbox{.}(2023)]%
        {yasunaga2023retrieval}
\bibfield{author}{\bibinfo{person}{Michihiro Yasunaga}, \bibinfo{person}{Armen
  Aghajanyan}, \bibinfo{person}{Weijia Shi}, \bibinfo{person}{Richard James},
  \bibinfo{person}{Jure Leskovec}, \bibinfo{person}{Percy Liang},
  \bibinfo{person}{Mike Lewis}, \bibinfo{person}{Luke Zettlemoyer}, {and}
  \bibinfo{person}{Wen-tau Yih}.} \bibinfo{year}{2023}\natexlab{}.
\newblock \showarticletitle{Retrieval-augmented multimodal language modeling}.
\newblock  (\bibinfo{year}{2023}).
\newblock


\bibitem[Ye et~al\mbox{.}(2023a)]%
        {ye2023mplugdoc}
\bibfield{author}{\bibinfo{person}{Jiabo Ye}, \bibinfo{person}{Anwen Hu},
  \bibinfo{person}{Haiyang Xu}, \bibinfo{person}{Qinghao Ye},
  \bibinfo{person}{Ming Yan}, \bibinfo{person}{Yuhao Dan},
  \bibinfo{person}{Chenlin Zhao}, \bibinfo{person}{Guohai Xu},
  \bibinfo{person}{Chenliang Li}, \bibinfo{person}{Junfeng Tian},
  {et~al\mbox{.}}} \bibinfo{year}{2023}\natexlab{a}.
\newblock \showarticletitle{mplug-docowl: Modularized multimodal large language
  model for document understanding}.
\newblock \bibinfo{journal}{\emph{arXiv preprint arXiv:2307.02499}}
  (\bibinfo{year}{2023}).
\newblock


\bibitem[Ye et~al\mbox{.}(2023b)]%
        {ye2023mplug}
\bibfield{author}{\bibinfo{person}{Qinghao Ye}, \bibinfo{person}{Haiyang Xu},
  \bibinfo{person}{Guohai Xu}, \bibinfo{person}{Jiabo Ye},
  \bibinfo{person}{Ming Yan}, \bibinfo{person}{Yiyang Zhou},
  \bibinfo{person}{Junyang Wang}, \bibinfo{person}{Anwen Hu},
  \bibinfo{person}{Pengcheng Shi}, \bibinfo{person}{Yaya Shi}, {et~al\mbox{.}}}
  \bibinfo{year}{2023}\natexlab{b}.
\newblock \showarticletitle{mplug-owl: Modularization empowers large language
  models with multimodality}.
\newblock \bibinfo{journal}{\emph{arXiv preprint arXiv:2304.14178}}
  (\bibinfo{year}{2023}).
\newblock


\bibitem[Yu et~al\mbox{.}(2023)]%
        {yu2023fusing}
\bibfield{author}{\bibinfo{person}{Youngjae Yu}, \bibinfo{person}{Jiwan Chung},
  \bibinfo{person}{Heeseung Yun}, \bibinfo{person}{Jack Hessel},
  \bibinfo{person}{Jae~Sung Park}, \bibinfo{person}{Ximing Lu},
  \bibinfo{person}{Rowan Zellers}, \bibinfo{person}{Prithviraj Ammanabrolu},
  \bibinfo{person}{Ronan Le~Bras}, \bibinfo{person}{Gunhee Kim}, {and}
  \bibinfo{person}{Yejin Choi}.} \bibinfo{year}{2023}\natexlab{}.
\newblock \showarticletitle{Fusing Pre-Trained Language Models With Multimodal
  Prompts Through Reinforcement Learning}. In
  \bibinfo{booktitle}{\emph{Proceedings of the IEEE/CVF Conference on Computer
  Vision and Pattern Recognition (CVPR)}}. \bibinfo{pages}{10845--10856}.
\newblock


\bibitem[Yuan et~al\mbox{.}(2021)]%
        {yuan2021florence}
\bibfield{author}{\bibinfo{person}{Lu Yuan}, \bibinfo{person}{Dongdong Chen},
  \bibinfo{person}{Yi-Ling Chen}, \bibinfo{person}{Noel Codella},
  \bibinfo{person}{Xiyang Dai}, \bibinfo{person}{Jianfeng Gao},
  \bibinfo{person}{Houdong Hu}, \bibinfo{person}{Xuedong Huang},
  \bibinfo{person}{Boxin Li}, \bibinfo{person}{Chunyuan Li}, {et~al\mbox{.}}}
  \bibinfo{year}{2021}\natexlab{}.
\newblock \showarticletitle{Florence: A new foundation model for computer
  vision}.
\newblock \bibinfo{journal}{\emph{arXiv preprint arXiv:2111.11432}}
  (\bibinfo{year}{2021}).
\newblock


\bibitem[Zhang et~al\mbox{.}(2023)]%
        {zhang2023meta}
\bibfield{author}{\bibinfo{person}{Yiyuan Zhang}, \bibinfo{person}{Kaixiong
  Gong}, \bibinfo{person}{Kaipeng Zhang}, \bibinfo{person}{Hongsheng Li},
  \bibinfo{person}{Yu Qiao}, \bibinfo{person}{Wanli Ouyang}, {and}
  \bibinfo{person}{Xiangyu Yue}.} \bibinfo{year}{2023}\natexlab{}.
\newblock \showarticletitle{Meta-transformer: A unified framework for
  multimodal learning}.
\newblock \bibinfo{journal}{\emph{arXiv preprint arXiv:2307.10802}}
  (\bibinfo{year}{2023}).
\newblock


\bibitem[Zhou et~al\mbox{.}(2023)]%
        {zhou2023comprehensive}
\bibfield{author}{\bibinfo{person}{Ce Zhou}, \bibinfo{person}{Qian Li},
  \bibinfo{person}{Chen Li}, \bibinfo{person}{Jun Yu}, \bibinfo{person}{Yixin
  Liu}, \bibinfo{person}{Guangjing Wang}, \bibinfo{person}{Kai Zhang},
  \bibinfo{person}{Cheng Ji}, \bibinfo{person}{Qiben Yan},
  \bibinfo{person}{Lifang He}, {et~al\mbox{.}}}
  \bibinfo{year}{2023}\natexlab{}.
\newblock \showarticletitle{A comprehensive survey on pretrained foundation
  models: A history from bert to chatgpt}.
\newblock \bibinfo{journal}{\emph{arXiv preprint arXiv:2302.09419}}
  (\bibinfo{year}{2023}).
\newblock


\bibitem[Zhou et~al\mbox{.}(2021)]%
        {zhou2021informer}
\bibfield{author}{\bibinfo{person}{Haoyi Zhou}, \bibinfo{person}{Shanghang
  Zhang}, \bibinfo{person}{Jieqi Peng}, \bibinfo{person}{Shuai Zhang},
  \bibinfo{person}{Jianxin Li}, \bibinfo{person}{Hui Xiong}, {and}
  \bibinfo{person}{Wancai Zhang}.} \bibinfo{year}{2021}\natexlab{}.
\newblock \showarticletitle{Informer: Beyond efficient transformer for long
  sequence time-series forecasting}. In \bibinfo{booktitle}{\emph{Proceedings
  of the AAAI conference on artificial intelligence}},
  Vol.~\bibinfo{volume}{35}. \bibinfo{pages}{11106--11115}.
\newblock


\bibitem[Zhu et~al\mbox{.}(2022)]%
        {zhu2022uni}
\bibfield{author}{\bibinfo{person}{Xizhou Zhu}, \bibinfo{person}{Jinguo Zhu},
  \bibinfo{person}{Hao Li}, \bibinfo{person}{Xiaoshi Wu},
  \bibinfo{person}{Hongsheng Li}, \bibinfo{person}{Xiaohua Wang}, {and}
  \bibinfo{person}{Jifeng Dai}.} \bibinfo{year}{2022}\natexlab{}.
\newblock \showarticletitle{Uni-perceiver: Pre-training unified architecture
  for generic perception for zero-shot and few-shot tasks}. In
  \bibinfo{booktitle}{\emph{Proceedings of the IEEE/CVF Conference on Computer
  Vision and Pattern Recognition}}. \bibinfo{pages}{16804--16815}.
\newblock


\end{thebibliography}

\appendix

\end{document}